\documentclass{amsart}
\usepackage{amsthm}
\usepackage{amsmath}
\usepackage{amsfonts}
\usepackage{amssymb}
\usepackage[hidelinks]{hyperref}
\usepackage[ruled,linesnumbered]{algorithm2e}
\usepackage[noend]{algpseudocode}
\usepackage{graphicx}
\usepackage{subfig}
\usepackage{chngcntr}
\counterwithin{figure}{section}
\counterwithin{table}{section}

\def\real{{\mathbb{R}}}

\def\Xs{\mathcal{X}}
\def\Ys{\mathcal{Y}}
\def\probS{\normalfont{\rm Prob}(\Xs \times \Ys)}
\def\planS{\normalfont{\rm Plan}(\mu, \nu)}
\def\tranS{\normalfont{\rm TM}(\mu, \nu)}
\def\potS{\normalfont{\rm Pot}(c)}
\def\discrS#1.{\normalfont{\rm disc }_{{\setbox0=\hbox{$#1\unskip$}\ifdim\wd0=0pt k
    \else #1\fi}}}
\def\symDiscrS#1.{\normalfont{\rm symdisc }_{{\setbox0=\hbox{$#1\unskip$}\ifdim\wd0=0pt k
    \else #1\fi}}}
\def\lagr{\mathcal{L}}
\def\ContS#1.{C({\setbox0=\hbox{$#1\unskip$}\ifdim\wd0=0pt \Xs\times\Ys
    \else #1\fi})}
\def\projX{\normalfont{\rm pr}_{\mathcal{\Xs}}}
\def\projY{\normalfont{\rm pr}_{\mathcal{\Ys}}}
\DeclareMathOperator*{\argmax}{arg\,max}
\DeclareMathOperator*{\argmin}{arg\,min}
\def\Wass#1.{{\normalfont{\rm W}}_{
\setbox0=\hbox{$#1\unskip$}\ifdim\wd0=0pt 0
    \else #1\fi}}
\begin{document}
\title[Learning to Transport]{\large Learning to Transport with Neural Networks \\
\small A comparison of mathematically sound \& heuristic approaches}
\author{Andrea Schioppa}
\address{Amsterdam, Noord Holland}
\email{ahisamuddatiirena+math@gmail.com}
\begin{abstract}
    We compare several approaches to learn an Optimal Map,
    represented as a neural network, between probability distributions.
    The approaches fall into two categories: ``Heuristics'' and
    approaches with a more sound mathematical justification,
    motivated by the dual of the Kantorovitch problem.
    Among the algorithms we consider a novel approach
    involving dynamic flows and reductions of Optimal Transport
    to supervised learning.
\end{abstract}
\maketitle
\tableofcontents
\section{Introduction}\label{sec:intro}
Many problems in Machine Learning require comparing
probability distributions.
Applications of computing distances between probability
histograms include bag-of-words for natural language processing~\cite{kusner-embeddings},
 color and shape processing in computer vision~\cite{solomon_conv_wasserstein},
and regularization of generative models~\cite{arjovsky_wgan, patrini_sink_auto}
\par While many distances between probability distributions have been
studied and used in the literature, the Optimal Transport distance~\cite{cuturi_book, santambrogio_book}
has come to hold a prominent place.
One main reason for its success is its flexibility as it can be
parametrized using a cost chosen depending on the application domain.
Another reason is that solving the Optimal Transport problem allows to
find a transformation, an Optimal Map, between two probability
distributions which minimizes a given loss function.
\par In particular, while computing a distance allows to say how
much two histograms are close to each other, computing a map
allows to solve additional problems, for example sampling from
intractable distributions using an optimal map to transform
a tractable distribution into an intractable one~\cite{trigila_thesis, seguy_neural_opt},
flow tracing in fluid dynamics and medical diagnostics~\cite{orlova_medical, trigila_thesis},
and
database aggregation between data collected in different laboratories~\cite{trigila_thesis}.
In my work in E-Commerce I have additionally used Optimal Transport to
design quasi-experiments~\cite{quasi_experiments_video_streaming, netflix_quasi_experiments}
in situations where adoption of a given product cannot be properly
randomized between treatment and control.
In many cases using Optimal Transport has allowed me to build
comparable treatment and control groups increasing the statistical
power to measure the effect of interventions.
\par Finally, there is increasing work in Optimal Transport in
general metrics spaces and with non-smooth costs~\cite{ambrosio_book, lott_rcd}
where the lack of regularity and high-dimensionality makes traditional
approaches based on the Monge-Ampere equation unfeasible.
\par In this work we make a comparison of different approaches to learn optimal
transport using Neural Networks.
We look at a data-driven formulation to solve the problem exploiting
a class of Machine Learning models that have gained prominence in the
last two decades with their successes in Natural Language Processing,
Computer Vision and Variational Inference.
\subsection{Previous work} \label{subsec:prev_work}
It his hard to make proper justice to the amount of literature
on computational Optimal Transport and its applications to Machine Learning.
We pick a few previous works related to this paper and refer the reader to~\cite{cuturi_book}.
In~\cite{trigila_thesis} Trigila and Tabak studied algorithms to find the
Optimal Map using dynamic flows.
Their approach is data-driven and motivated by problems where the
probability histograms live in high-dimensional spaces.
\par In~\cite{cuturi_lightspeed} Cuturi showed that Optimal Transport distances, after
adding an entropic regularization term, can be computed
efficiently using the celebrated Sinkhorn's algorithm~\cite{sinkhorn1967}.
The work~\cite{cuturi_lightspeed} is a landmark in making optimal transportation
distances attractive to the Machine Learning community.
Moreover, in~\cite{genevay_large_opt} the authors showed how to
scale up the computation of distances using online learning and
stochastic gradient descent.
\par Optimal transport distances have soon attracted the interest of
people working on Generative Models as the optimal transport distance
can be used as a regularization term.
For example in~\cite{arjovsky_wgan} the authors introduced Wasserstein GANs,
which are GANs where the adversarial network computes an $l_1$-optimal
transport distance, and show the superior performance of Wasserstein GANs
over traditional GANs.
Moreover they also provided a mathematical justification for preferring
optimal transport to other regularization terms, a line of research
further developed in~\cite{arjovsky_principled}.
Another recent work using the Sinkhorn algorithm is the paper~\cite{patrini_sink_auto}
introducing the so-called Sinkhorn autoencoders.
\par The first work we are aware of that uses Neural Network to
compute transport maps is~\cite{seguy_neural_opt} where the authors show
applications to generative models and domain adaptation.
\subsection{Contributions} \label{subsec:contributions}
The contribution of this paper is mainly comparing several
approaches to learn the optimal transport map using neural networks.
From the existing literature we take two approaches employed in~\cite{seguy_neural_opt},
namely solving the dual transport problem with
entropic / $l_2$-regularization.
Novel approaches that we consider are using neural networks
in dynamic flows motivated by~\cite{trigila_thesis}, adversarial training
for the optimal map motivated by
the literature on generative models~\cite{arjovsky_wgan, goodfellow_gans, lecun_energy_wgans},
and supervised learning, which is motivated by the assumption that
discrete / semi-discrete problems can be used to train
statistical models which are capable of generalization.
\subsection{Code Links} \label{subsec:code_links}
We try to make this work reproducible providing the code and data in the
GitHub repository:\hfill\null\linebreak
\href{https://github.com/salayatana66/learn_to_transport_code}{https://github.com/salayatana66/learn\_to\_transport\_code}.
In the following we just refer to this repository as the \texttt{GitRepo}.
\par Please note that the experiments/code there is not an exhaustive representation
of the experiments we tried out.
Quite a bit of work went into engineering each experiment and decide which parameters to test.
This was broken down into intermediate experiments that resulted in the final scripts.
Concretely, in Subsection~\ref{subsec:learn_tr_metric_spaces} we discuss the importance of
proper initialization.
We learned this while experimenting on learning linear maps between Gaussians in different dimension,
code which we exclude from the \texttt{GitRepo}.
Similarly, in Subsection~\ref{subsec:learn_tr_super_learn} we discuss the issue of normalization
while learning potentials in a supervised way.
Making that work took a bit of effort, also in light of numerical instability of the Sinkhorn iterations
for small values of the regularization.
The final solution is based on going to log-space, compare~\cite[Subsec.~4.4]{cuturi_book}.
\section{Review of Optimal Transport} \label{sec:opt_review}
\subsection{Notation} \label{subsec:opt_review_not}
We now recall the formulation of the Optimal Transport Problem.
Let $\Xs$ and $\Ys$ denote two metric spaces and let $\probS$ denote the
  set of Radon probability measures on $\Xs\times\Ys$, and
for each metric space $\mathcal{Z}$ let $\ContS \mathcal{Z}.$ denote the
set of real valued continuous functions defined on $\mathcal{Z}$.
Let $\projX$, $\projY$ denote the projections of $\Xs\times\Ys$ on the
  first and second factor, respectively.
In the following we fix a probability measure $\mu$ on $\Xs$ and another one
  $\nu$ on $\Ys$.
In the optimal transport problem we want to move $\mu$ into $\nu$
  minimizing some kind of cost;
  therefore, to each pair of points $(x,y)\in\Xs\times\Ys$ we associate
  a \textbf{transport cost} $c(x,y)$, and, without loss of generality, we
  can assume that $c\in\ContS.$ and $c$ is non-negative.
\subsection{The Kantorovich Formulation} \label{subsec:opt_review_kan}
  Let $\planS$ denote the set of those $\pi\in\probS$ such that the marginals
  on $\Xs$ and $\Ys$ give back $\mu$, $\nu$: ${\projX}_{\#}\pi=\mu$,
  ${\projY}_{\#}\pi=\nu$.
   In the \textbf{Kantorovich formulation} the
  optimal transport problem becomes:
  \begin{equation}\label{eq:kant_primal}
  \argmin_{\pi}\left\{\int c(x,y)d\pi(x,y): \pi\in\planS\right\}\tag{$\mathcal{P}_0$};
  \end{equation}
  the infimum value of the objective in \eqref{eq:kant_primal} will
  be denoted by $\Wass.$.
  We will refer to problem \eqref{eq:kant_primal} as the \textbf{un-regularized primal}.
  \par In cases of common interest $\Xs=\Ys=\real^d$ and the cost $c$ is the
  square of the Euclidean distance; in that case $\Wass.^{1/2}$ is
  called \textbf{$l_2$-Wasserstein distance} between the probability distributions
  $\mu$ and $\nu$.
\subsection{The Monge formulation} \label{subsec:opt_review_monge}
  Having found a minimizer $\pi$ for \eqref{eq:kant_primal} one can
  use the Disintegration Theorem~\cite[Box~2.2]{santambrogio_book} to represent $\pi$ as:
  \begin{equation}\label{eq:disintegration}
  \pi = \int_{\Xs}d\mu(x)\int_{\projY^{-1}(x)}\pi_x,
  \end{equation}
  where $x\mapsto\pi_x$ is a map associating to each $x\in\Xs$ in the support
  of $\mu$ a family of probability measures in $\probS$ supported on the fiber
  of $x$ under the projection.
  The probability $\pi_x$ can be interpreted as a transport plan for a unit
  of mass at $x$ towards points of $Y$ in the support of $\nu$.
  In the \textbf{Monge formulation} of the transport problem one requires
  that $\pi_x$ is concentrated at a single point $T(x)$ of $\Ys$; equivalently
  there is a (measurable )map $T:\Xs\to\Ys$ (note there is need to define $T$
  only on the support of $\Xs$) such that
  $({\normalfont{\rm Id}}, T)_{\#}\mu=\pi$.
  One can also directly formulate the Monge Problem: let $\tranS$ denote the set of
  those measurable maps $T:\Xs\to\Ys$ with $T_{\#}\mu=\nu$; then the problem becomes:
  \begin{equation}\label{eq:monge}
  \argmin_{T}\left\{\int c(T(x),x)d\mu(x): T\in\tranS\right\}\tag{$\mathcal{M}_0$}.
  \end{equation}
  The convexity and linearity of \eqref{eq:kant_primal} implies the existence of a minimizer $\pi$
  under mild conditions~\cite[Theorem~1.4]{santambrogio_book}.
  On the other hand in the Monge Problem it is not always the case that $\tranS$
  is nonempty or that the infima of the objectives of \eqref{eq:kant_primal} and
  \eqref{eq:monge} are the same;
 we refer the reader to~\cite[Secs~1.3--1.5]{santambrogio_book} for more information.
  A map $T\in\tranS$ is a \textbf{transport map} and an \textbf{optimal transport map}
  is a solution to \eqref{eq:monge}.
  \par In this paper we are mainly concerned with finding an approximation to an optimal $T$
  and we will deliberately neglect issues about the existence and regularity of such a $T$.
  We just observe that given a plan $\pi\in\probS$ and mild assumptions on $c(x,y)$ (e.g.~convexity or being a distance)
  there is a simple heuristic to associate to
  it a transport map: for each $x$ in the support of $\mu$ one defines $T(x)$ as
  \begin{equation}\label{eq:pi_to_T_heur}
  \argmin_{\tilde y}\left\{\int_{\projY^{-1}(x)} c(\tilde y, y)d\pi_x(y): \tilde y\in\Ys\right\}\tag{$\pi\xrightarrow{\textrm{heur}} T$};
  \end{equation}
  for example in the case of the quadratic Euclidean cost one maps $x$ to the barycenter of
  $\pi_x$.
  \subsection{The dual problem} \label{subsec:opt_review_dual}
  The problem \eqref{eq:kant_primal} is a convex problem;
  its dual can be formulated in terms of a pair of potentials $u\in\ContS\Xs.$,
  $v\in\ContS\Ys.$.
  We let $\potS$ denote the set of those pairs $(u,v)\in\ContS\Xs.\times\ContS\Ys.$
  such that for each $(x,y)$ one has $u(x)+v(y)\le c(x,y)$. The dual of \eqref{eq:kant_primal} is then:
  \begin{equation}\label{eq:kant_dual}
    \argmax_{u, v}\left\{\int u\,d\mu + \int v\,d\nu: (u,v)\in\potS\right\}\tag{$\mathcal{D}_0$}.
  \end{equation}
  One might show~\cite[Theorems~1.37,1.38]{santambrogio_book} that given an optimal $(u,v)$ for \eqref{eq:kant_dual} an
  optimal $\pi$ for \eqref{eq:kant_primal} can be found using the disintegration formula:
  $\pi_x$ can be taken as the uniform probability distribution on those $y$ such that
  $u(x) + v(y) = c(x,y)$.
    \subsection{Regularization} \label{subsec:opt_review_reg}
    The problem \eqref{eq:kant_primal} is not strongly convex;
    adding a regularization term it can be turned into a strongly convex one
    improving its convergence properties.
    Let $\mu\otimes\nu\in\probS$ denote the
    product of $\mu$ and $\nu$ and for $\pi\in\planS$ let
    $\frac{d\pi}{d\mu\otimes\nu}$ be the Radon-Nikodym of $\pi$ with respect
    to $\mu\otimes\nu$.
    If $\pi$ is not absolutely continuous with respect
    to $\mu\otimes\nu$, in Real Analysis it is common practice to restrict
    the Radon Nikodym derivative to the part of $\pi$ which is absolutely
    continuous with respect to $\mu\otimes\nu$.
    However in Optimal Transport it is a common practice to set the derivative to $+\infty$ in
    such a case.
    As we will then consider regularizations which are $+\infty$ in the case
    of $\pi$ not being absolutely continuous with respect to $\mu\otimes\nu$, optimal plans
    for \eqref{eq:kant_primal_ereg} will
    be in general ``diffused'' or ``smoothed'' by the regularization.
    In fact, if the optimal plan $\pi$ for \eqref{eq:kant_primal} is induced by an optimal
    transport map $T$ for \eqref{eq:monge}, then $\pi$ is generally singular with respect to $\mu\otimes\nu$.
    \par For $\varepsilon >0$ the \textbf{entropic regularization} of \eqref{eq:kant_primal} is
    \begin{equation}
      \label{eq:kant_primal_ereg}
      \begin{split}
     \argmin_{\pi}\bigg\{\int c(x,y)\,d\pi(x,y)\\
    &+\varepsilon\int\left(
    \log\frac{d\pi}{d\mu\otimes\nu} - 1
    \right)\,d\pi(x,y): \pi\in\planS\bigg\}.
   \end{split} \tag{$\mathcal{P}_\varepsilon$}
  \end{equation}
  The dual of \eqref{eq:kant_primal_ereg} is the \textbf{unconstrained} problem:
  \begin{equation}\label{eq:kant_dual_ereg}
  \begin{split}
  \argmax_{u, v}\bigg\{\int  u\,d\mu + \int v\,d\nu
  &-\varepsilon\int\exp\left(
  \frac{u(x)+v(y)-c(x,y)}{\varepsilon}
  \right)\,d\mu\otimes\nu(x,y):\\ &\quad (u,v)\in\ContS\Xs.\times
  \ContS\Ys.\bigg\}.
  \end{split}\tag{$\mathcal{D}_\varepsilon$}
  \end{equation}
  Given a solution $(u,v)$ for \eqref{eq:kant_dual_ereg} on obtains a solution
  $\pi$ for \eqref{eq:kant_primal_ereg} setting:
  \begin{equation}\label{eq:rockafellar_ereg}
  d\pi(x,y) = \exp\left(\frac{u(x)+v(y)-c(x,y)}{\varepsilon}\right)\,
    d\mu\otimes\nu(x,y).
  \end{equation}
  \par An alternative regularization is the $l_2$-one which gives the problems:
  \begin{equation}
      \label{eq:kant_primal_l2reg}
      \begin{split}
     \argmin_{\pi}\bigg\{\int c(x,y)\,d\pi(x,y)
    &+\varepsilon\int\left(
    \frac{d\pi}{d\mu\otimes\nu}
    \right)^2\,d\pi(x,y):\\ &\quad \pi\in\planS\bigg\},
   \end{split} \tag{$\mathcal{P}^{l_2}_\varepsilon$}
  \end{equation}
  \begin{equation}\label{eq:kant_dual_l2reg}
  \begin{split}
  \argmax_{u, v}\bigg\{\int & u\,d\mu + \int v\,d\nu\\
  &-\frac{1}{4\varepsilon}\int
(u(x)+v(y)-c(x,y))_{+}^2\,d\mu\otimes\nu(x,y):
  \quad(u,v)\in\ContS\Xs.\times
  \ContS\Ys.\bigg\};
  \end{split}\tag{$\mathcal{D}^{l_2}_\varepsilon$}
  \end{equation}
  and the analogue of \eqref{eq:rockafellar_ereg} becomes
  \begin{equation}\label{eq:rockafellar_l2reg}
  d\pi(x,y) = \frac{(u(x)+v(y)-c(x,y))_{+}}{2\varepsilon}\,
    d\mu\otimes\nu(x,y).
  \end{equation}
  \subsection{Discretization} \label{subsec:opt_review_disc}
  When $\mu$ and $\nu$ are discrete measures all problems \eqref{eq:kant_primal},
  \eqref{eq:monge}, \eqref{eq:kant_dual} and their regularized versions become
  discrete optimization problems.
  For problem \eqref{eq:kant_dual_ereg} there is an efficient computation
  algorithm, the \textbf{Sinkhorn algorithm}~\cite{sinkhorn1967}, with a convergence rate
  $O(\varepsilon^{-3}\log n)$, $n$ being an upper bound on the number of points
  at which $\mu$, $\nu$ are supported.
  The Sinkhorn algorithm is a matrix rescaling procedure that can be
  efficiently parallelized on GPUs~\cite{cuturi_lightspeed} and in the limit $\varepsilon\to 0$ one
  can guarantee to recover a solution of the unregularized problem.
  However, in practice too small values of $\varepsilon$ lead to rounding
  errors and slow convergence;
  here we have not used values smaller than $0.005$.
  \section{Learning to Transport} \label{sec:learn_tr}
\begin{figure}[ht]
        \caption{Different approaches to learn an Optimal Map}
        \label{fig:approaches}
    \includegraphics[height=5cm,width=12cm]{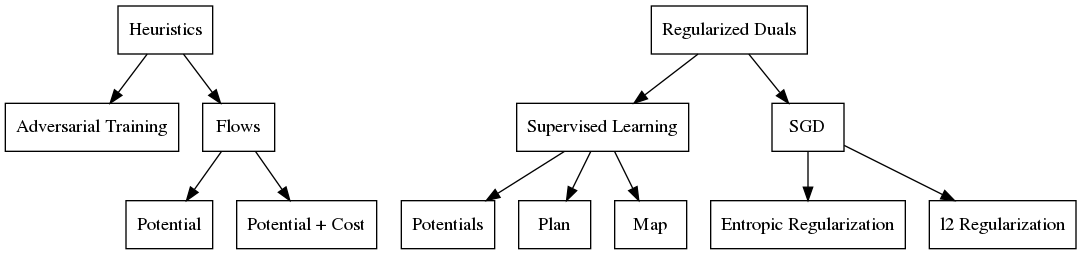}
    \end{figure}

  Figure~\ref{fig:approaches} summarizes the approaches that we examine to learn a transport map $T$ from $\mu$
  to $\nu$ where $T$ is represented by a Neural Network.
  We can classify the approaches in two categories:
  \begin{enumerate}
    \item \textbf{Heuristics}: the mathematical justification is not so solid, but the approach
    is motivated by intuition and might work reasonably well in practice. In the Machine Learning
    literature it is common to resort to such approaches when problems are hard to treat or
    correctly formulate (e.g.~Variational Inference, Variational AutoEncoders, Generative Adversarial
    Neural Networks).
    \item \textbf{Solving a regularized dual}: these are based on solving \eqref{eq:kant_dual_ereg} or
    \eqref{eq:kant_dual_l2reg}.
    If $\mu$ and $\nu$ are discrete the theory is classical;
    when $\mu$ or $\nu$ are continuous and the potentials $u$ and $v$ are represented via neural
    networks we do not know of rigorous convergence guarantees, but methods can work well in practice.
  \end{enumerate}
    \subsection{Representations} \label{subsec:learn_tr_repr}
    What is a good way to represent mathematical objects like $\mu$, $\nu$, $T$, $u$ or $v$?
    The answer will likely depend on the setting.
    For example $\mu$ and $\nu$ can be discretized on a fixed mesh and then one needs to keep
    track of the value of maps and functions only on a fixed set of points.
    As observed in~\cite{trigila_thesis} mesh-based approaches do not scale well in the dimension and
    for Big Data applications~\cite{trigila_thesis} advocates having access to random samples drawn from $\mu$ and
    $\nu$.
    For a discussion of discretization and Wasserstein distances look at~\cite{weed_convergence}.
    \par From a Machine Learning perspective it is also convenient to store $T$ as a model,
    hence in a compact form.
    Besides \textbf{storage efficiency}, one also gains in \textbf{generalization} power.
    For example assume taht $T$ has been learned on discrete samples;
    now such a $T$ can be also evaluated on new samples.
    Here we focus on representing $T$ (or $u$, $v$ or $\pi$ depending on the circumstances)
    as neural networks.
    The first reason is that even with just two hidden layers one can represent
    rich spaces of non linear functions.
    The second reason is availability of Open Source libraries to train Neural Networks, here we will
    use PyTorch.
    \subsection{Flows} \label{subsec:learn_tr_flows}
     A first set of heuristics to find a transport map $T$ considers \textbf{flows}, i.e.~dynamically
     evolving trajectories from points sampled from $\mu$ to points sampled from $\nu$.
     We loosely follow~\cite{trigila_thesis} to outline the main idea and, just for the moment, we let $\Xs$, $\Ys$
     be subsets of some Euclidean space $\real^d$ and assume that $\mu$ and $\nu$ are continuous
     with respect to the Lebesgue measure.
     In this case \eqref{eq:monge} an \eqref{eq:kant_primal} will have the same infimum value and
     there will be an optimal map inducing an optimal plan.
     In~\cite{trigila_thesis} the authors try to reach the optimal $T$ using a flow $\{T_t\}_t$ starting
     at the identity.
     Under the aforementioned regularity assumptions they are able to combine the
     Monge and Kantorovich formulations by updating at each step $T_t$ and the potentials $u$ and $v$.
     This is made possible by linearizing the Monge Ampere equation with different approaches
     that they investigate.
     To fix the ideas, a simple linearization approach consists in using a linear perturbation of the
     identity (equivalently assuming that $u$ and $v$ are updated via second order polynomials),
     which result in $\{T_t\}$ being represented as a composition of linear maps.
     A drawback of this approach is that to compute $T$ at new unseen points one needs to unroll
     the chain of compositions.
     \par Therefore, here we propose to parametrize $T$ as a neural network $T_w$, where $w$
     are the weights initialized at a point $w_0$ such that $T_{w_0}$ is the identity
     map.
     Having defined a Lagrangian $\lagr(T_{w(t)}, \mu, \nu)$ we can simply evolve $w(t)$
     by gradient descent:
     \begin{equation}\label{eq:w_sgd_lagr}
     \frac{dw(t)}{dt} = -\nabla_{w(t)}\lagr(T_{w(t)}, \mu, \nu).
     \end{equation}
     In reality updates are done stochastically drawing batches of samples from $\mu$ and $\nu$.
     Note however, that while we still use flows, we are no longer making regularity assumptions
     on the measure $\mu$, $\nu$ or the spaces $\Xs$ and $\Ys$.
  \subsection{Flow Potentials} \label{subsec:learn_tr_flow_pot}
   Let us first forget about the cost part and look for flows that would induce $T_{w(t)\#}\mu\to\nu$
  as $t\to\infty$.
  Here we look for Lagrangians of the form $\lagr(T_{w(t)\#}\mu, \nu)$.
  If one had $T_{w(t)\#}\mu = \nu$ then for each $f\in\ContS\Ys.$ one would have:
  \begin{equation}
    \int_{\Xs}f(T_{w(t)}(x))\,d\mu(x) = \int_{\Ys}f(y)\,d\nu(y).
  \end{equation}
  Therefore if we choose a sufficiently rich family $\{f_k\}_{k=1}^K\subset\ContS\Ys.$ we might
  hope of flowing $\mu$ into $\nu$ by using a Lagrangian of the form:
  \begin{equation}\label{eq:lagr_base}
    \lagr(T_{w(t)\#}\mu, \nu) = \sum_{k=1}^K\left|
    \int_{\Xs}f_k(T_{w(t)}(x))\,d\mu(x) - \int_{\Ys}f_k(y)\,d\nu(y)\right|^2.
  \end{equation}
   In reality, at each gradient descent iteration, one has access to batches
   $\{X_i\}_{i=1}^{b_X}\sim\mu$, $\{Y_j\}_{j=1}^{b_Y}\sim\nu$ and the Lagrangian becomes:
  \begin{equation}
    \lagr(\{T_{w(t)}(X_i)\}_{i=1}^{b_X}, \{Y_j\}_{j=1}^{b_Y}) = \sum_{k=1}^K\left|
    \frac{1}{b_X}\sum_{i=1}^{b_X}f_k(T_{w(t)}(X_i)) -
    \frac{1}{b_Y}\sum_{j=1}^{b_Y}f_k(Y_j)\right|^2.
  \end{equation}
  Motivated by~\cite{trigila_thesis} in our experiments we choose quadratic polynomials
  to match means and covariances, or concentrated Gaussians with centers lying on a grid
  $\{z_k\}_{k=1}^K$:
  \begin{equation}
    f_k(y)=\exp\left(\frac{-d(y, z_k)^2}{\sigma^2}\right).
  \end{equation}
  \par For a more geometric approach we could try to directly match the ``shapes'' of
  $\{X_i\}_{i=1}^{b_X}$, $\{Y_j\}_{j=1}^{b_Y}$ in a Hausdorff sense~\cite[Subsec~10.6.1]{cuturi_book}.
  Specifically assume that $\Xs$ and $\Ys$ are subsets of a metric space $\mathcal{Z}$ with distance
  $d$.
  Fix $t$ and let $Y_{j_l(i)}$ denote an $l$-th closest point of $\{Y_j\}_{j=1}^{b_Y}$
  to $T_{w(t)}(X_i)$ (i.e.~sort the points in ascending order of distance from $T_{w(t)}(X_i)$,
breaking ties arbitrarily, and take the $l$-th point).
  Similarly let $T_{w(t)}(X_{i_l(j)})$ denote an $l$-th closest point of
 $\{T_{w(t)}(X_i)\}_{i=1}^{b_X}$ to $Y_j$.
  We can then define a \textbf{discrepancy @ k} and a \textbf{symmetric discrepancy @ k}:
  \begin{align}
    \discrS.(\{T_{w(t)}(X_i)\}_{i=1}^{b_X}, \{Y_j\}_{j=1}^{b_Y}) &= \frac{1}{b_X}
    \sum_{i=1}^{b_X}\sum_{l\le k}d(T_{w(t)}(X_i), Y_{j_l(i)}) \tag{discrepancy @ k}\\
    \symDiscrS.(\{T_{w(t)}(X_i)\}_{i=1}^{b_X}, \{Y_j\}_{j=1}^{b_Y}) &= \discrS.(\{T_{w(t)}(X_i)\}_{i=1}^{b_X}, \{Y_j\}_{j=1}^{b_Y})
    + \frac{1}{b_Y}\sum_{j=1}^{b_Y}\sum_{l\le k}d(T_{w(t)}(X_{i_l(j)}), Y_{j}).
    \tag{symmetric discrepancy @ k}
  \end{align}
  Note that $\discrS.$ will tend to move the mass of $\mu$ inside the support of $\nu$;
  while $\symDiscrS.$ will tend to move the mass of $\mu$ inside the support of $\nu$ while
  matching the shape of $\nu$, thus avoiding collapsing $\mu$ to a measure singular with
 respect to $\nu$.
  \par Finally, to bring back the optimal transport problem \eqref{eq:monge} we simply
  introduce a term depending on $c$ which acts as a \textbf{regularization} for the flow
  associated with $\lagr(T_{w(t)\#}\mu, \nu)$:
  \begin{equation}\label{eq:tp_cost_reg}
     \lagr(T_{w(t)}, \mu, \nu) = \int_{\Xs}c(x, T_{w(t)}(x))\,d\mu(x) +
    \lagr(T_{w(t)\#}\mu, \nu).
  \end{equation}
  \subsection{Adversarial Training} \label{subsec:learn_tr_adverse}
    \textbf{Adversarial training} formulates the learning problem as a min-max:
    \begin{equation}\label{eq:adverse_min_max}
        \min_{w}\max_{\theta}\bigg[
                \int_{\Xs}c(T_w(x), x)\,d\mu(x) + \int_{\Xs}f_{\theta}(T_w(x))\,d\mu(x)
                - \int_{\Ys}f_{\theta}(y)\,d\nu(y)
    \bigg]\tag{$Adv$}
    \end{equation}
    where the neural network $\theta\mapsto f_{\theta}$ parametrizes a function in $\ContS\Ys.$ to
    penalize choices of $w$ such that $T_{w\,\#}\mu\ne\nu$.
    In a typical experiment in Euclidean space one would initialize $T_w$ to the identity
    and $f_\theta$ to the zero function.
    \par The theoretical justification for \eqref{eq:adverse_min_max} is that the inner $\max$
    over $\theta$
    would give $+\infty$ if $T_{w\,\#}\mu\ne\nu$.
    As in the case of GANs~\cite{goodfellow_gans} in practice one cannot really train a min-max because of the
    vanishing gradients problem~\cite{arjovsky_principled}.
    In a typical application for each minimization step updating $w$ there is a loop of a fixed
    number of steps updating $\theta$.
    \par In~\cite{arjovsky_wgan} it is observed that if one could force $f_\theta$ to be $L$-Lipschitz for some
    $L>0$, then the penalization term in \eqref{eq:adverse_min_max} involving $f_\theta$ would
    converge to a multiple of the $l_1$-Wasserstein distance between the measures $T_{w\,\#}\mu$
    and $\nu$.
    In WGANs the authors propose a heuristic to keep $f_\theta$ Lipschitz by using gradient clipping.
    In our experiments we did not benefit much from gradient clipping and we are not quite
    sure whether the min-max training would yield a penalization term close to a Wasserstein distance.
  \subsection{Regularized duals after~\cite{seguy_neural_opt}} \label{subsec:learn_tr_reg_dual_seguy}
   In~\cite{seguy_neural_opt} the authors propose to parametrize $u$ and $v$ in \eqref{eq:kant_dual_ereg}
    or \eqref{eq:kant_dual_l2reg} as neural networks $u_\theta$, $v_\eta$.
    They use stochastic gradient descent to learn an optimal $(u_{\theta^*},v_{\eta^*})$
    and can thus generate a transport plan using \eqref{eq:rockafellar_ereg} or
    \eqref{eq:rockafellar_l2reg}.
    Finally they also parametrize the transport map $T$ as a neural network $T_w$
    and use the heuristic \eqref{eq:pi_to_T_heur} to learn the parameter $w$.
    \par Despite lack of rigorous guarantees on optimizing the regularized dual(s) using neural networks, this approach
    solves scalability issues on large data-sets.
    Assume that $\mu$ (resp.~$\nu$) is represented by storing observed data
    in a data-set $\mathbf{X}$ (resp.~$\mathbf{Y}$).
    Then the cost $c$ can be stored as a matrix which grows as $\#\mathbf{X}\times\#\mathbf{Y}$,
    making its storage cost prohibitive.
    The advantage of~\cite{seguy_neural_opt} is that one needs only to compute $c$
    on each batch, incurring a cost growing as $b_X\times b_Y$.
    Nevertheless, it might be the case that using heuristics like in~\cite{maggioni_package} one might be able to sparsfiy $c$ reducing its storage cost.
    However, it seems like that scaling~\cite{maggioni_package} to large datasets
    would require distributing the computation across several nodes in
    a cluster.
    \par To summarize, the approach of~\cite{seguy_neural_opt} is very effective as large
    data-sets can be streamed to a single machine incurring a fixed cost in
    memory usage.
    Note that the dimensionality of the parameters $\theta$, $\eta$, $w$
    will grow with the dimensions of the spaces $\Xs$, $\Ys$ so there is still an
    effect of the dimensionality on the scalability of this approach.
    Moreover, it might also be the case that in high dimensions large batch sizes
    might be needed to properly train the neural networks~\cite{weed_convergence}.
    We think investigation of the interrelation between converge,
    batch sizes and dimensionality of $\Xs$, $\Ys$
    is an interesting topic for further research.
    \subsection{Supervised learning} \label{subsec:learn_tr_super_learn}
    An alternative approach to~\cite{seguy_neural_opt} is to use \eqref{eq:kant_dual_ereg}
    as a ``source of truth'' to generate a stream of training data.
    One would then train the neural networks using this data, reducing
    optimal transport to a supervised learning task.
    \par To fix the ideas, let us go back to learning $u_\theta$, $v_\eta$;
    on each batch we can use the Sinkhorn Algorithm~\cite{sinkhorn1967} to find
    optimal $\{\hat u_i\}_{i=1}^{b_X}$, $\{\hat v_j\}_{j=1}^{b_Y}$ and
    then iterate to minimize a loss like:
    \begin{equation}\label{eq:sink_iters1}
        \sum_{i=1}^{b_X}|u_\theta(X_i) - \hat u_i| +
        \sum_{j=1}^{b_Y}|v_\eta(Y_j) - \hat v_j|.
    \end{equation}
    An advantage of this approach is that Sinkhorn iterations, especially
    on small batches, are much faster to converge than stochastic gradient
    descent for \eqref{eq:kant_dual_ereg}.
    Moreover, learning the source of truth for large data-sets
    might be carried out in parallel on different nodes in a cluster;
    finally, the training data would be streamed to a single node
    to optimize the parameters $\theta$, $\eta$.
    \par In the case in which the training data is still fitted at the
    time of training $u_\theta$, $v_\eta$, a further advantage of this
    approach is that one can supply $\{u_\theta(X_i)\}_{i=1}^{b_X}$,
    $\{v_\eta(X_j)\}_{j=1}^{b_Y}$ as starting values for the Sinkhorn
    iterations.
    A disadvantage of this approach involving potentials is that
    each solution $\{\hat u_i\}_{i=1}^{b_X}$, $\{\hat v_j\}_{j=1}^{b_Y}$
    is only well-defined up to an additive constant, i.e.~$\{\hat u_i -C\}_{i=1}^{b_X}$, $\{\hat v_j +C\}_{j=1}^{b_Y}$
    would still yield a solution.
    In our experiments we overcome this issue enforcing some kind of
    normalization.
    \par There is nothing special about learning $u_\theta$, $v_\eta$:
    one can apply this approach to directly solve on the batches for
    an optimal plan $\hat\pi_{i,j}$ or transport map $\hat T$ and then
    train neural networks $\pi_{w}$ or $T_{w}$ to approximate $\hat\pi_{i,j}$,
    $\hat T$.
    In our experiments we also explore these approaches.
    Learning $\hat T$ is particularly favorable as one avoids
    the additional step in~\cite{seguy_neural_opt} of using the potentials to learn
    the map.
    \par Finally, in this approach there is nothing special about
    Sinkhorn iterations or \eqref{eq:kant_dual_ereg}.
    One can use the reduction of optimal transport to supervised
    learning anytime there is a good black-box approach to learn
    potentials, plans or maps on the minibatches.
    \subsection{Initializations and general metric spaces} \label{subsec:learn_tr_metric_spaces}
    In our experiments we focus on the cases in which $\Xs$ and $\Ys$
    are subsets of $\real^d$ and $c$ is the standard quadratic cost.
    In that case we initialize potentials to $0$, plans to the
    product measure and maps to the identity.
    At the start of the project we made simple tests on using
    neural networks to learn a transport map which is linear.
    We found that even in $1$-dimensions initialization to the
    identity can be important.
    For example, if $T_{w_0}$ has negative determinant, e.g.~in the
    case in which $T_{w_0}$ reverses some space directions,
    we find that during training updates to the parameters
    $w_0$ will tend to keep the determinant negative.
    \par In the case of $\Xs$, $\Ys$ being general metric spaces (e.g.~Carnot groups or graphs)
    one can always initialize potentials to $0$ or plans to the product measure.
    However, initialization / representation of the transport map will depend on
    the metric space.
    We think this is also an interesting area for further research.
   \section{Experiments} \label{sec:experiments}
For details on the commands we used to run the experiments we
refer to the \texttt{GitRepo}, in file \texttt{Experiments.md}.
Performance metrics and pictures can be found in \texttt{evaluation\_metrics/}
and \texttt{snapshot\_images\_and\_movies/}.
    \subsection{The Dataset} \label{subsec:data}
    The crucial decision we have taken at the beginning of our work
    has been to limit our tests to a \emph{single} dataset while
    exploring a \emph{variety} of algorithms.
    We have deliberately decided to keep the dataset as simple as possible;
    however we have avoided trying to learn a linear transport map.
    \par
\begin{figure}[ht]
        \caption{Visualization of $\mu$, $\nu$ with 1024 samples}
        \label{fig:dataset}
    \includegraphics[height=8cm,width=10cm]{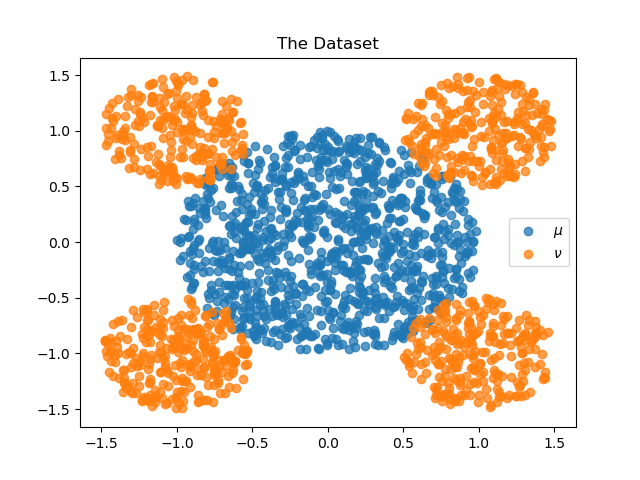}
    \end{figure}

 We have thus opted for planar measures where the transport map is not
    smooth as it requires splitting the domain of $\mu$ into pieces.
    Specifically, $\mu$ is the uniform distribution on the unit ball;
    $\nu$ is a uniform distribution supported on $4$ balls of radius
    $\frac{1}{2}$ and centers at the points $(\pm 1, \pm 1)$.
    See Figure~\ref{fig:dataset} for a visualization.

    \subsection{Performance} \label{subsec:perf}
    In order to assess the performance of each algorithm we compute the mean
    squared distance between the transport map $T_w$ and the ``optimal'' one
    $T_{\textrm{opt}}$:
\begin{equation}
    \varepsilon_2 = \int_{\Xs} \|T_w(x)-T_{\textrm{opt}}\|_2^2\,d\mu(x).
\end{equation}
    To define $T_{\textrm{opt}}$ we take samples  $\{X_i\}_{i=1}^{B}\sim\mu$,
    $\{Y_j\}_{j=1}^B\sim\nu$ ($B=1000$ in our experiments) and compute
    a ``ground truth'' $T_{\textrm{opt}}$ using the Sinkhorn algorithm
    with a low value of the entropic regularization (i.e.~we set $\varepsilon=10^{-2}$
    in \eqref{eq:kant_primal_ereg} in our experiments).
We thus compute:
\begin{equation}
    \varepsilon_2=\frac{1}{B}\sum_{i=1}^B\|T_w(X_i) - T_{\textrm{opt}}(X_i)\|^2.
\end{equation}
We stress that our experiment are \textbf{dynamic}, i.e.~$T_w$ is the
final point of a ``flow'' $\{T_{w(t)}\}_t$ starting at $T_{w(0)}$ being
the identity map (at least when restricted to the support of $\mu$).
We thus take a fixed number $S$ ($S=50$ in our experiments) of snapshots of
$T_{w(t)}$ across the training iterations $t\in\{0,\cdots,T-1\}$.
In this way we are both able to estimate the \textbf{rate of convergence}
and the \textbf{stability} of the training process.
\par A non trivial thing to account for is that different algorithms
run for a different number $T$ of iterations.
In our plots we account for this normalizing the steps with $t\mapsto t/T$
to compare the convergence rate across the iterations.
Once we single out promising algorithms we are able to dig more
into the training times using statistics that we log at periodic intervals
in Tensorboard.
\par
\begin{figure}[ht]
        \caption{Example of a ``movie frame'' generated during training.
        Blue points are sampled from $\mu$, and black ones from $\nu$.
        Red points are the images of each blue point under $T_{w(t)}$ and
        we keep them linked to their ``source point'' via a black segment.}
        \label{fig:example_movie_frame}
    \includegraphics[height=8cm,width=10cm]{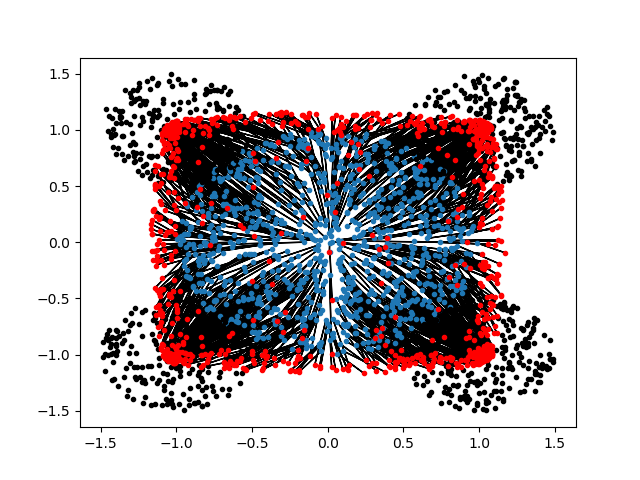}
    \end{figure}

 Finally, for each of the $S$ timesteps we save an image of the map
$T_{w(s)}$, see for example figure~\ref{fig:example_movie_frame}.
At the end of training we compose the images into a movie (see \texttt{GitRepo})
to get an idea of the flow and also inspect visually the ``quality'' of
the final map.
\par Even though visual inspection is not as objective as using an
evaluation metric, it allows us to compare maps that have similar
error $\varepsilon_2$ but different properties.
For example we will see cases where $\varepsilon_2$ is similar
between two different algorithms but in once case the final map
is too much diffused around the support to $\nu$, while in the
second case the final map is squashed inside the support of $\nu$.
We will see also cases where $\varepsilon_2$ is comparable to the one of
a map far from the optimal transport map, but nevertheless on
visual inspection the final map looks quite ``reasonable''.
\par While writing the training scripts we found qualitative methods
helpful in debugging errors.
Indeed at the start of this project we were just
looking at metrics in Tensorboard, but quickly
realized that we were not getting enough insights into the behavior of
the ``flow'' $\{T_{w(t)}\}_t$.
Obviously we are also helped by our choice of a lower dimensional dataset.
\par We think a possible area of further research is to design metrics
that make these objective inspections more quantitative.
We find that in practice just looking at the transport cost is not enough.
Good inspection metrics should take into account not just how far
or close $T_{w(t)\#}\mu$ is from the support of $\nu$, but also
the relative shapes of the measures.
\subsection{Comparison of Heuristics} \label{subsec:heur_comp}
Here is a summary of experiments we ran using the heuristics in
Subsection~\ref{subsec:learn_tr_flow_pot}:
\begin{itemize}
    \item \textit{covariance}: We try to force matching of second order
    momenta.
    \item \textit{exp}: We use a grid of centers and use a family of
    Gaussian bumps in~\eqref{eq:lagr_base}.
    \item \textit{discr\_N}: We use \textbf{discrepancy @ k}.
    \item \textit{sym\_discr\_N}: We use \textbf{symmetric discrepancy @ k}.
\end{itemize}
When we add \textit{tp\_} in front of an experiment name we use the
cost as a regularization term, compare~\eqref{eq:tp_cost_reg}.
\par Results are reported in Table~\ref{table:heur_opt_perf};
$\varepsilon_2$ is the minimum $\varepsilon_2$ across the $S$ snapshot
iterations;
the $\varepsilon_2$ is realized at iteration $t_{\textrm{min}}$ out
of the total $T$ iterations.
The standard deviation $\sigma(\varepsilon_2)$ of $\varepsilon_2$ is
computed on the iterations in the $S$ snapshots that occur after
$t_{\textrm{min}}$.
\begin{table}[ht]
\centering
\caption{Performance of ``heuristic'' flows}\label{table:heur_opt_perf}
\begin{tabular}{r|r|r|r|r}
  \hline
 model name & $\varepsilon_2$ & $\sigma(\varepsilon_2)$ &
  $t_{\textrm{min}}$ & $T$ \\
  \hline
covariance & 0.44 & 0.010 & 1000 & 5000 \\
  discr\_1 & 0.76 & 0.124 & 2000 & 100000 \\
  discr\_5 & 0.63 & 0.183 & 800 & 40000 \\
  exp & 0.31 & 0.019 & 9000 & 30000 \\
  sym\_discr\_1 & 0.28 & 0.023 & 21000 & 50000 \\
  sym\_discr\_5 & 0.26 & 0.039 & 21000 & 50000 \\
  tp\_covariance & 0.44 & 0.012 & 1000 & 5000 \\
  tp\_discr\_1 & 0.46 & 0.020 & 7000 & 50000 \\
  tp\_discr\_5 & 0.46 & 0.017 & 5400 & 30000 \\
  tp\_exp & 0.17 & 0.013 & 27000 & 30000 \\
  tp\_sym\_discr\_1 & 0.29 & 0.014 & 36000 & 50000 \\
  tp\_sym\_discr\_5 & 0.23 & 0.020 & 23000 & 50000 \\
   \hline
\end{tabular}
\end{table}
    \par
\begin{figure}[ht]
        \caption{The final transport yielded by \textit{tp\_covariance}}
        \label{fig:final_covariance}
    \includegraphics[height=8cm,width=10cm]{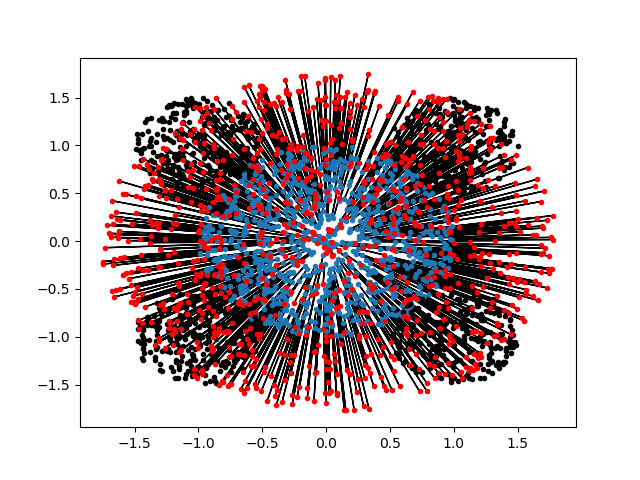}
    \end{figure}

\begin{figure}[ht]
        \caption{Convergence rate for \textit{covariance} and \textit{tp\_covariance}}
        \includegraphics[height=7cm,width=12cm]{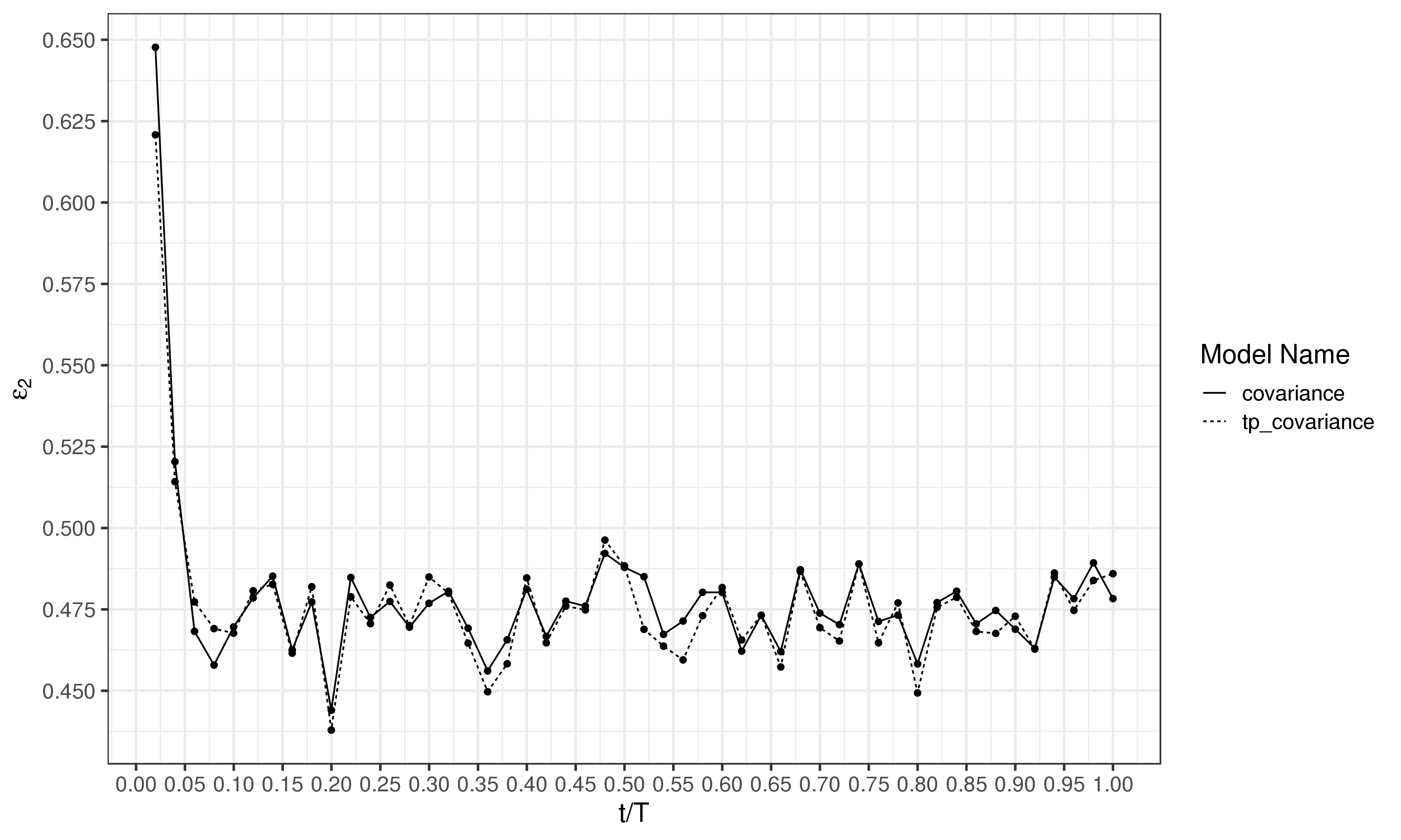}
                \label{fig:covariance_convergence}

    \end{figure}

 Except for \textit{covariance} all methods benefit by
    adding the cost regularization.
    The final transport yielded by \textit{covariance} is quite
    diffused, see Figure~\ref{fig:final_covariance}, and does not
    improve substantially after
    10\% of the iterations, see Figure~\ref{fig:covariance_convergence}.
\par
\begin{figure}[ht]
        \caption{Convergence rate for \textit{discr\_1} and \textit{discr\_5}}
        \includegraphics[height=7cm,width=12cm]{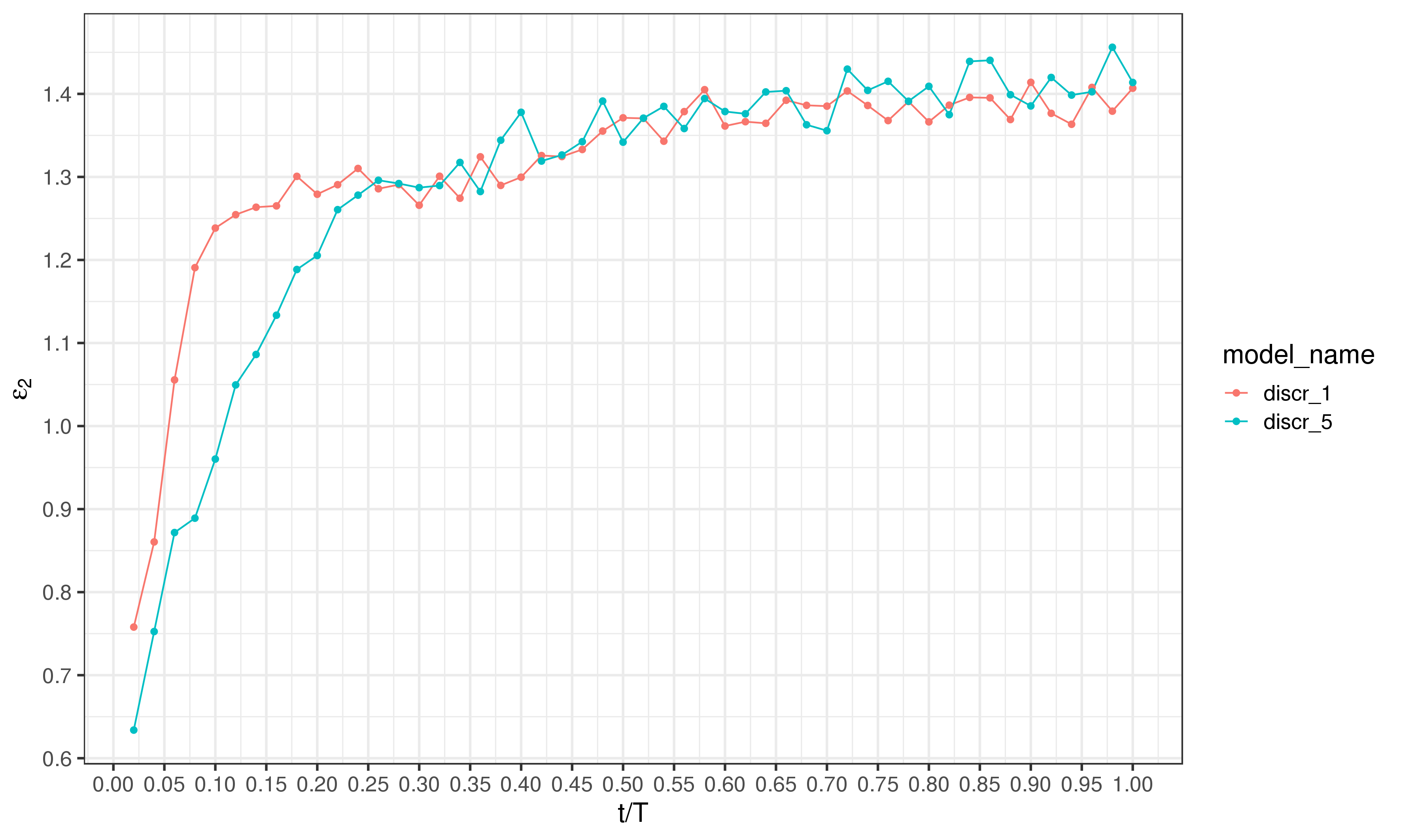}
                \label{fig:discr_convergence}
    \end{figure}

\begin{figure}[ht]
        \caption{Collapse of transport map for \textit{discr\_5}, $t=8000$}
        \includegraphics[height=8cm,width=10cm]{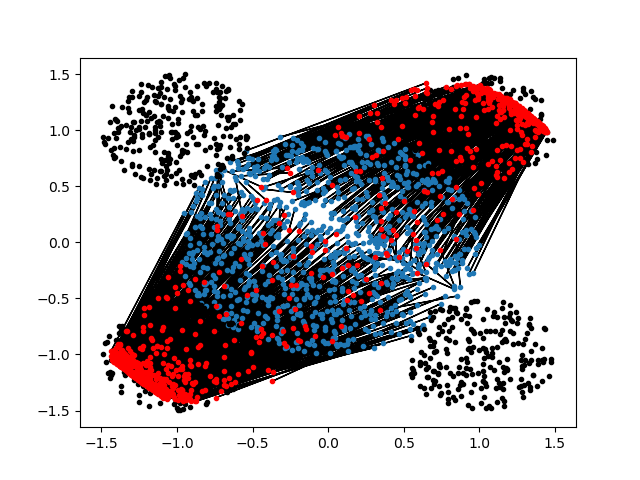}
                \label{fig:discr_collapse}
    \end{figure}

\begin{figure}[ht]
        \caption{Convergence rate for \textit{tp\_discr\_1} and
        \textit{tp\_discr\_5}}
        \includegraphics[height=7cm,width=12cm]{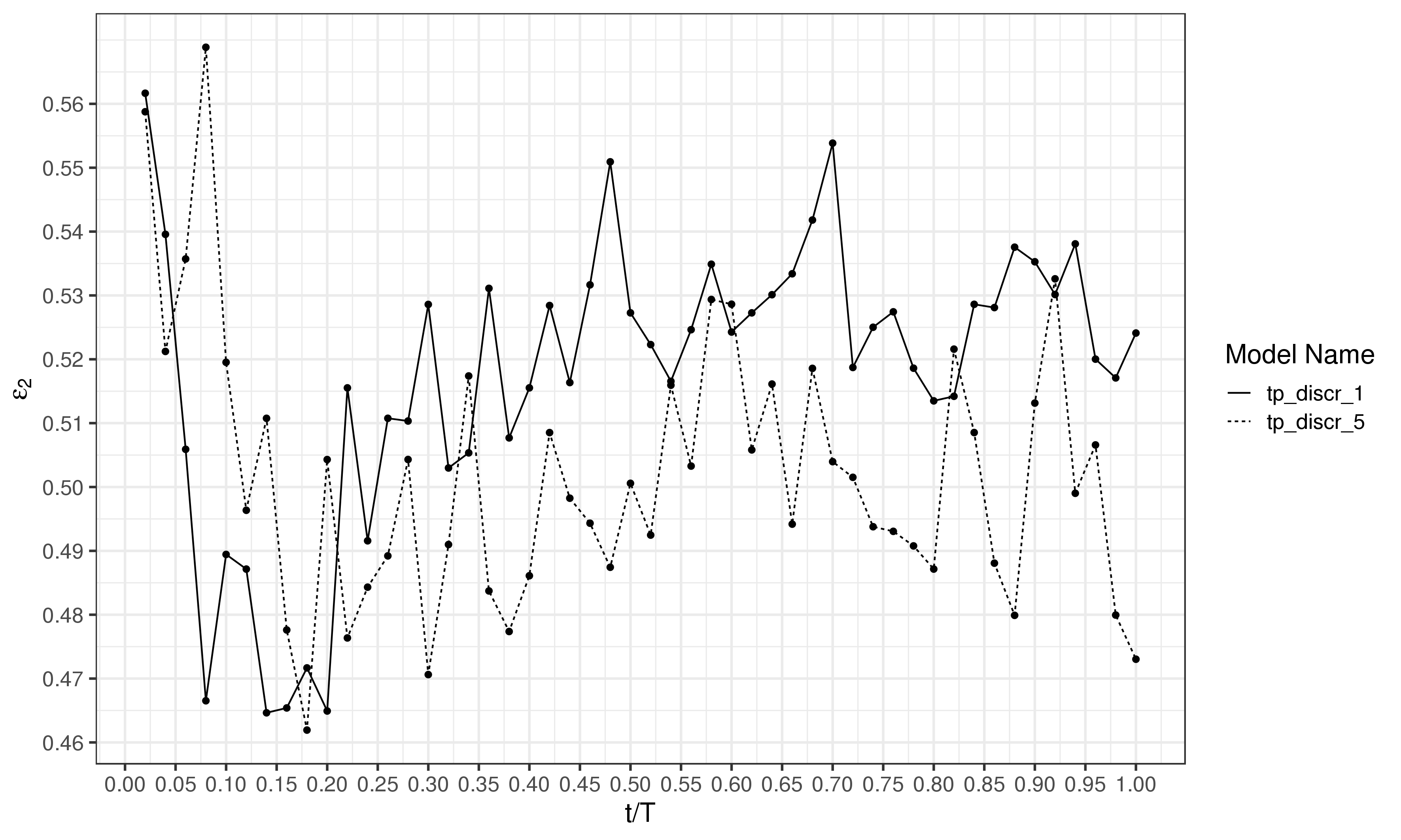}
                \label{fig:discr_tp_convergence}
    \end{figure}

\begin{figure}[ht]
        \caption{Collapse of transport map for \textit{tp\_discr\_5}, $t=7800$}
        \includegraphics[height=8cm,width=10cm]{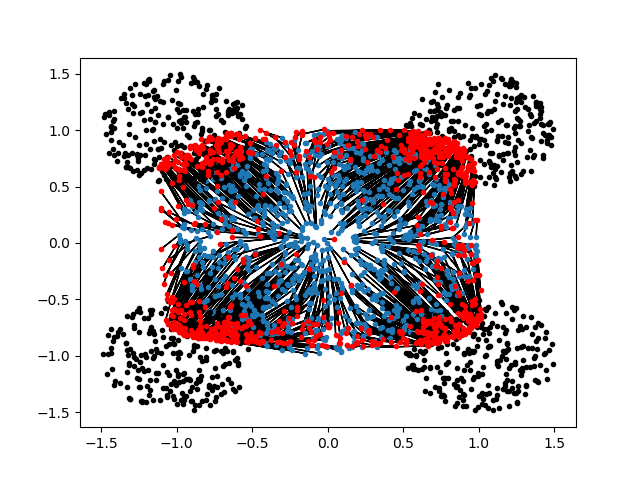}
                \label{fig:discr_tp_collapse}
    \end{figure}

 For the flows \textit{discr\_N} the error $\varepsilon_2$ increases
    with the number of iterations, see Figure~\ref{fig:discr_convergence}.
    We observe a quick chaotic push to collapse
    $T_{\#}\mu$ inside the support of $\nu$, see Figure~\ref{fig:discr_collapse}.
    Adding regularization improves these flows but the results remain
    disappointing, not improving over \textit{covariance} and not
    making substantial progress after the first 20\% of the iterations, see
    Figure~\ref{fig:discr_tp_convergence}.
    Despite adding regularization we still observe a sharp collapsing tendency,
see Figure~\ref{fig:discr_tp_collapse}.
\par
\begin{figure}[ht]
        \caption{Convergence for the flows based on \textbf{symdiscr @ } $N$}
        \includegraphics[height=7cm,width=12cm]{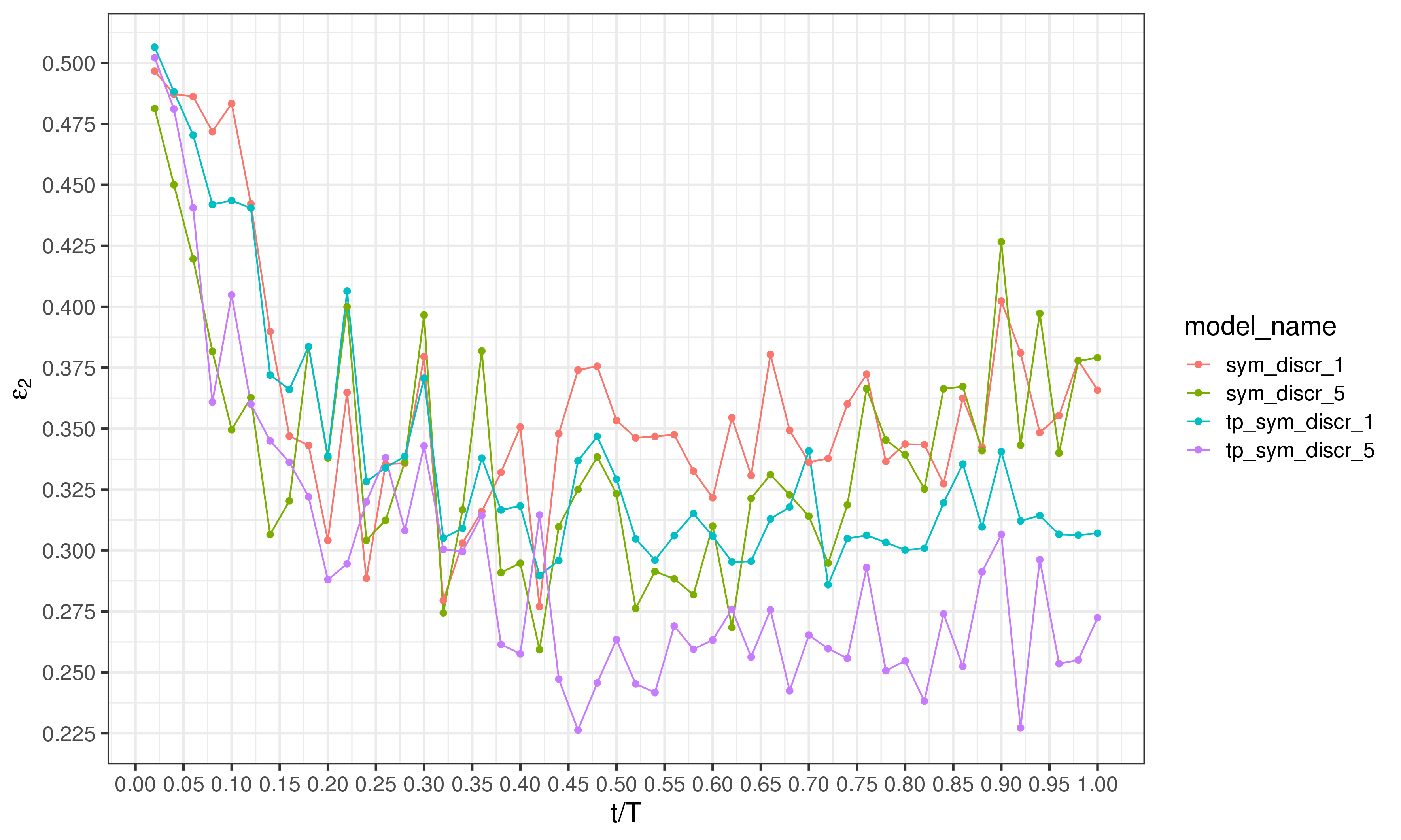}
                \label{fig:symdiscr_convergence}
    \end{figure}

\begin{figure}[ht]
        \caption{Best transport map for \textit{tp\_sym\_discr\_5}, $t=23000$}
        \includegraphics[height=8cm,width=10cm]{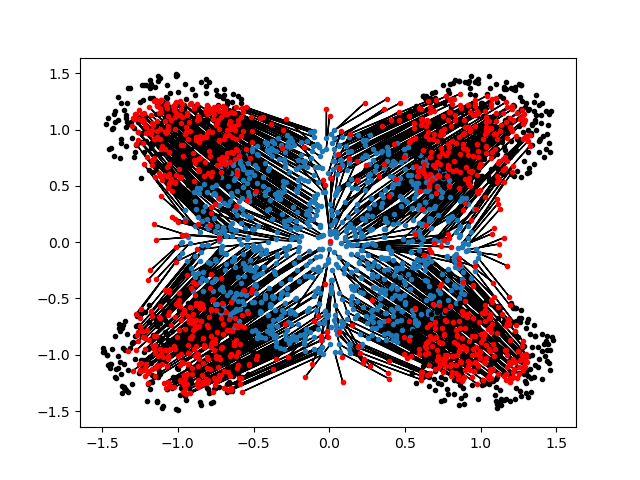}
                \label{fig:tp_symdiscr_converged}
    \end{figure}

 The flow \textit{sym\_discr} benefits a little bit by adding regularization
    with \textit{tp\_sym\_discr\_5} reaching the minimal $\varepsilon_2$.
    After the first 50\% of iterations there is not a real
    improvement, see Figure~\ref{fig:symdiscr_convergence}.
    On visual inspection this flow is quite successful in mapping
    $\mu$ into the support of $\nu$ while keeping the shapes of
    $T_{\#}\mu$ and $\nu$ comparable, see Figure~\ref{fig:tp_symdiscr_converged}.
    \par
\begin{figure}[ht]
        \caption{Convergence for the flows \textit{exp}, \textit{tp\_exp}}
        \includegraphics[height=7cm,width=12cm]{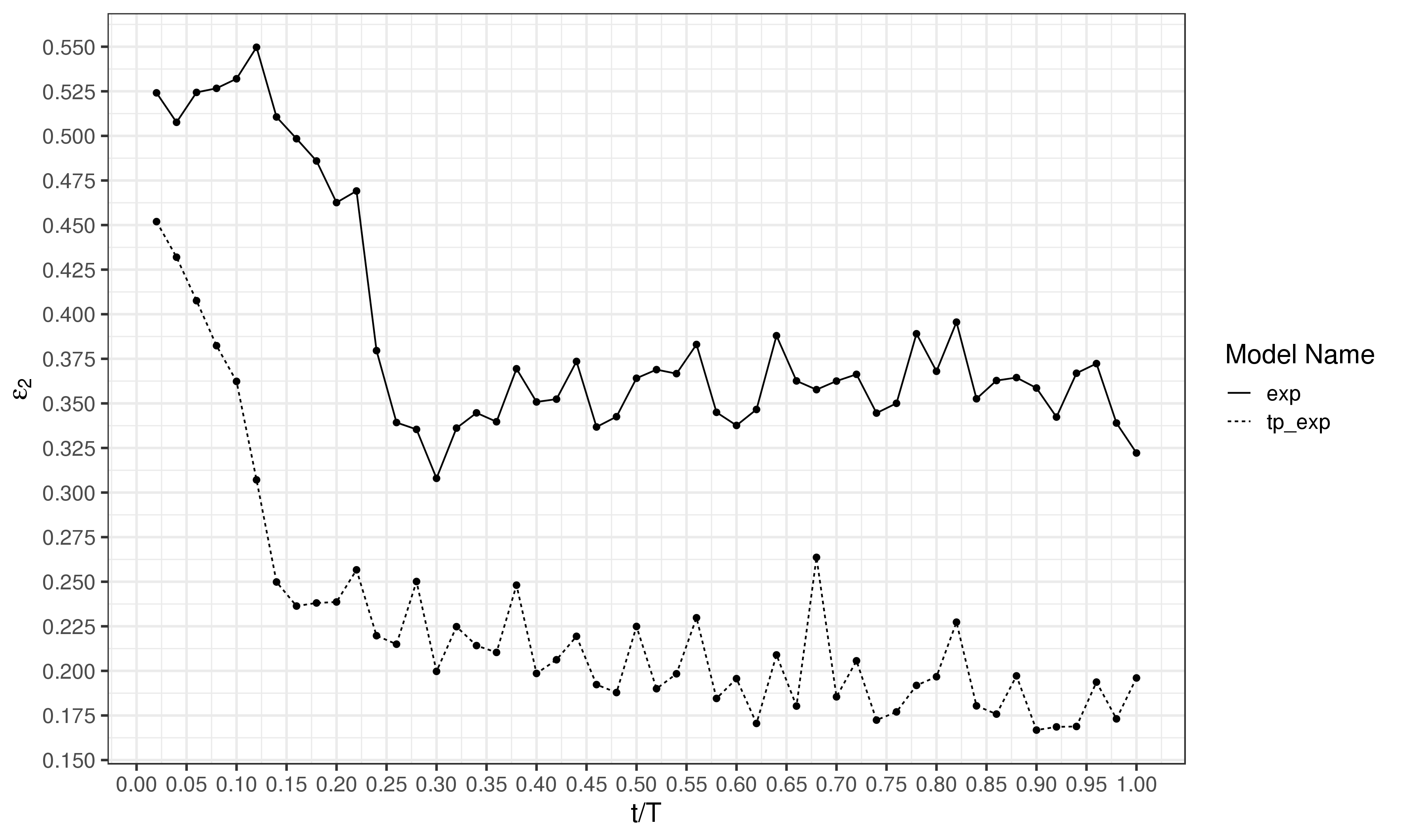}
                \label{fig:exp_convergence}
    \end{figure}

\begin{figure}[ht]
        \caption{Transport map for \textit{exp}, $t=9000$}
        \includegraphics[height=8cm,width=10cm]{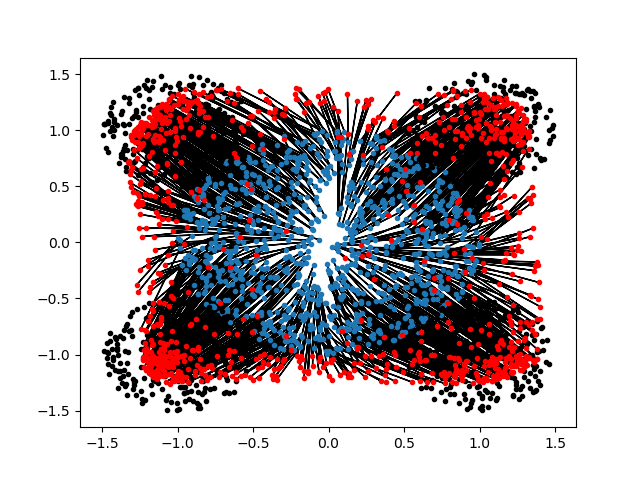}
                \label{fig:exp_map_frame}
    \end{figure}

\begin{figure}[ht]
        \caption{Transport map for \textit{tp\_exp}, $t=5400$}
        \includegraphics[height=8cm,width=10cm]{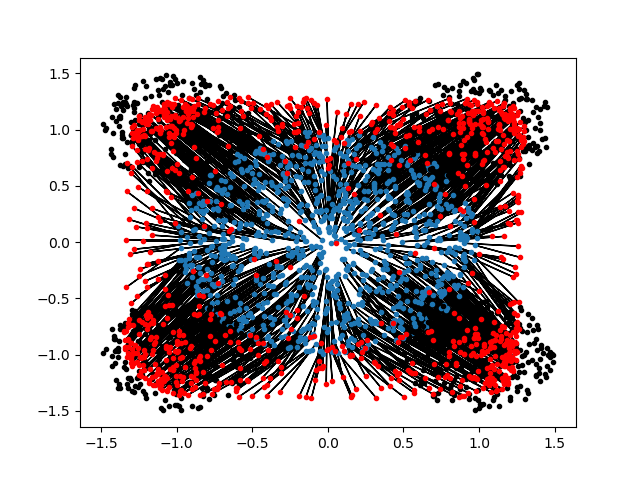}
                \label{fig:tp_exp_map_frame}
    \end{figure}

 The flows based on \textit{exp} give the best results and greatly
    benefit from adding the regularization.
The algorithm \textit{exp} reaches the minimum after about 30\% of the
    iterations while \textit{tp\_exp} does not improve substantially
after 60\% of the iterations, see Figure~\ref{fig:exp_convergence}.
    Both flows are quick to map $\mu$ into a measure matching the geometry of
    $\nu$, see Figures~\ref{fig:exp_map_frame} and \ref{fig:tp_exp_map_frame}.
\subsection{Comparison of Adversarial Models} \label{subsec:adv_comp}
Table~\ref{table:adv_models} summarizes the results of trying to learn
a transport map following the ``adversarial'' approach of Subsection~\ref{subsec:learn_tr_adverse}.
The name of each model refers to training parameters: for example
naming a model \textit{adv\_l10.0\_3\_clip\_0.01} means that the adversarial
term (i.e.~the one involving integration over $f_\theta$) in~\eqref{eq:adverse_min_max}
is ``boosted'' by multiplying $f_\theta$ by a factor $\lambda = 10$;
moreover the ``3'' means that for each gradient descent step for the parameters $w$,
2 gradient ascent steps are performed for the parameters $\theta$;
finally if ``clip'' is used, one is specifying the
parameter to clip the gradients, e.g.~in our case to $0.01$.
\par
\begin{figure}[ht]
        \caption{Convergence for the ``adversarial'' models}
        \includegraphics[height=7cm,width=12cm]{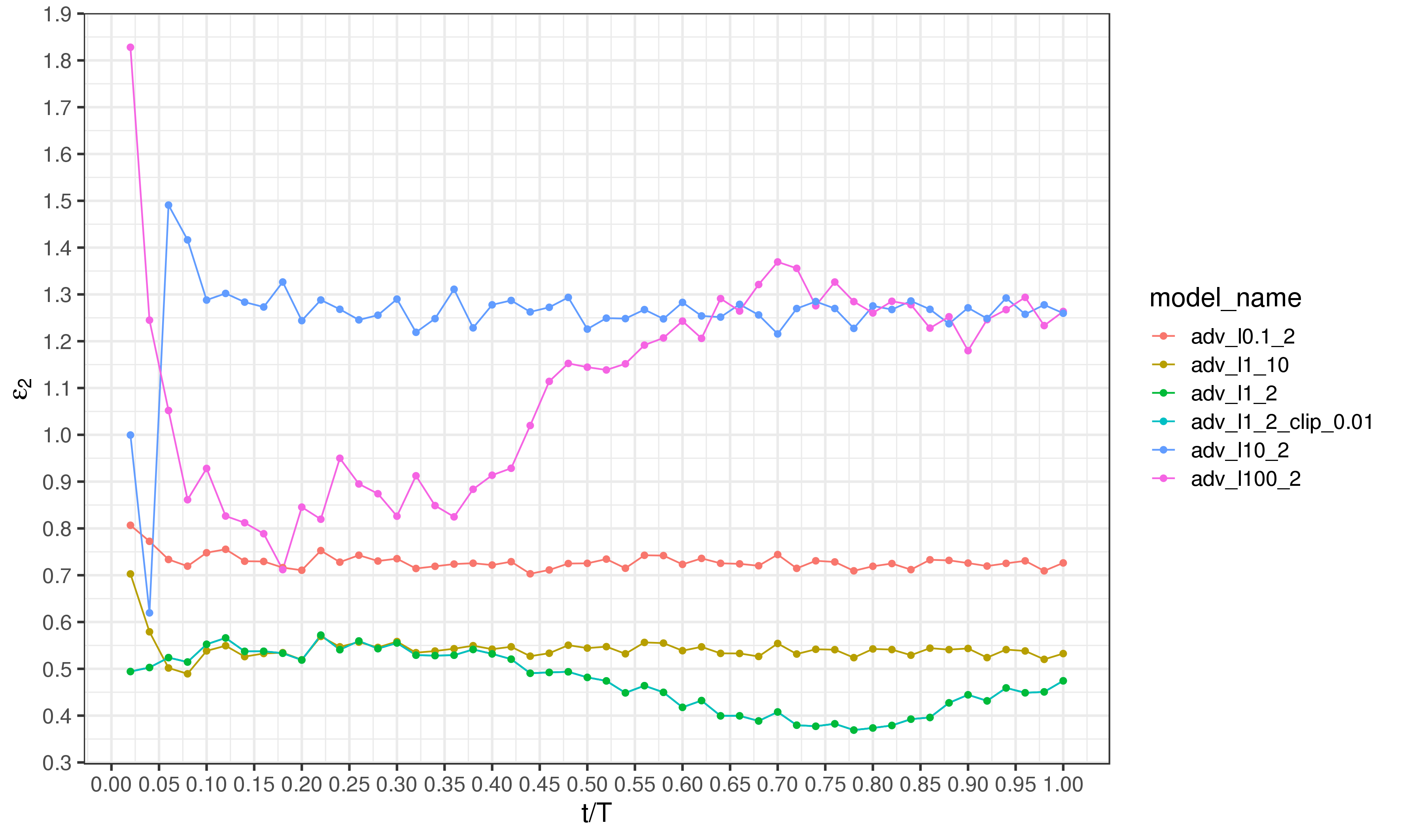}
                \label{fig:adversarial_convergence}
    \end{figure}

\begin{figure}
    \centering
    \caption{Chaotic behavior of adversarial training for $\lambda=100$
    at iterations $t=1000,8000,36000$ (from left to right).}
                    \label{fig:adv_large_lambda}
    \subfloat{\includegraphics[width=4cm]{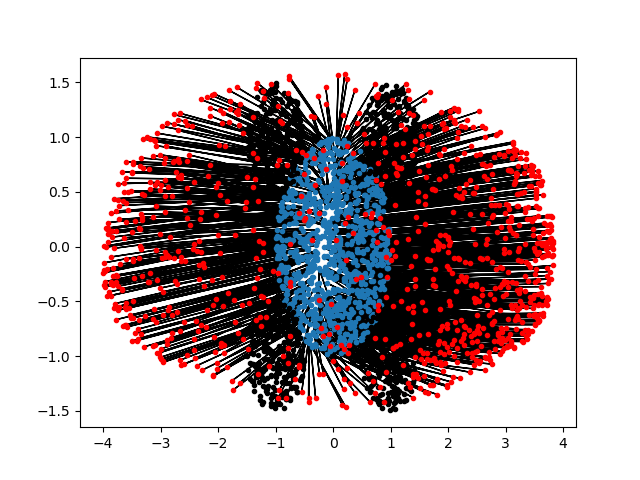}}
   \subfloat{\includegraphics[width=4cm]{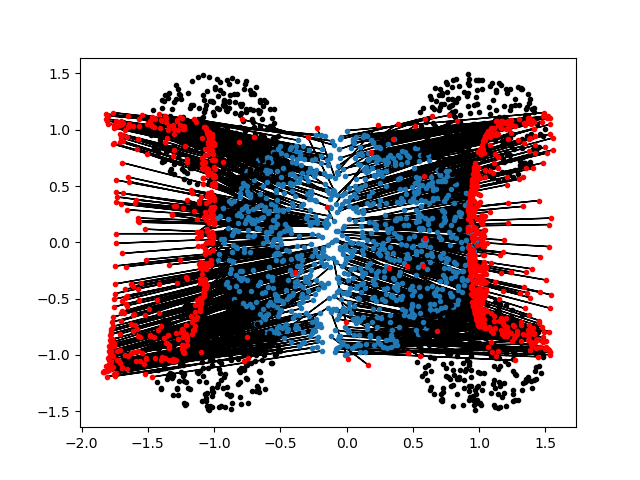}}
    \subfloat{\includegraphics[width=4cm]{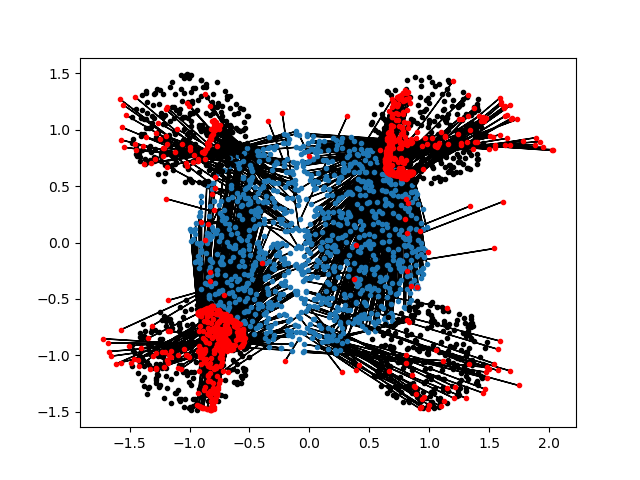}}
\end{figure}

 In Figure~\ref{fig:adversarial_convergence} we can inspect
the convergence rate of the adversarial models.
We see that the ``out-of-the-box'' parameter $\lambda=1$ seems
to yield the best results and gradient clipping does not seems to help.
Large values of $\lambda$ make the training unstable;
for example with $\lambda=100$ there is a steady increase in $\varepsilon_2$
after about 25\% of the training iterations.
Inspection of frames shows a lot of chaotic behavior in this case,
compare Figure~\ref{fig:adv_large_lambda}.
\par
\begin{figure}[ht]
        \caption{Transport map for \textit{adv\_l1\_2}, $t=39000$}
        \includegraphics[height=8cm,width=10cm]{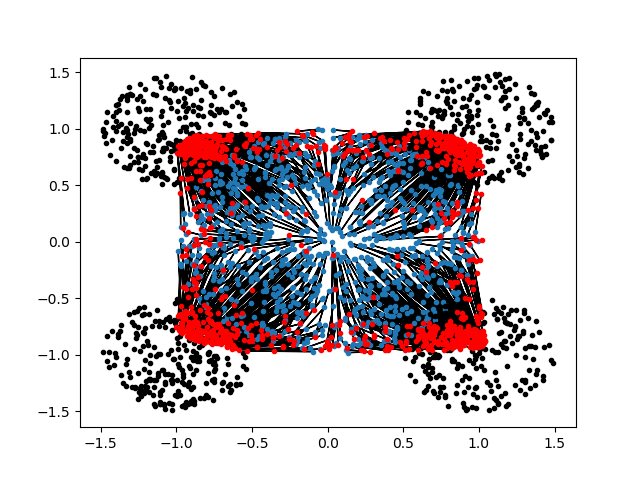}
                \label{fig:adv_frame_inspection}
    \end{figure}

 Also increasing the number of iterations of the adversarial
network leaves it stuck at a point with a higher value of $\varepsilon_2$.
The best model is \textit{adv\_l1\_2} which reaches the minimal
$\varepsilon_2$ at around 75\% of the iterations.
However, the final results are quite disappointing, both
in terms of $\varepsilon_2$, which is higher than the one
obtained simply with the \textit{exp} flow, and of the final map as
$T_{w(T)\#}\mu$ lies into a subset of the support of $\nu$, see
Figure~\ref{fig:adv_frame_inspection}.
Generally speaking, we found adversarial training to be under-performing
and very sensitive to parameter specification.
\par
\begin{figure}[ht]
        \caption{Comparison of the cost and adversarial component of the loss
        for \textit{adv\_l1\_2} across the training steps. Note we use
        statistics in TensorBoard so we are not restricted to the $S$
        snapshots.}
                        \label{fig:adversarial_cost_adv_losses}
        \includegraphics[height=7cm,width=12cm]{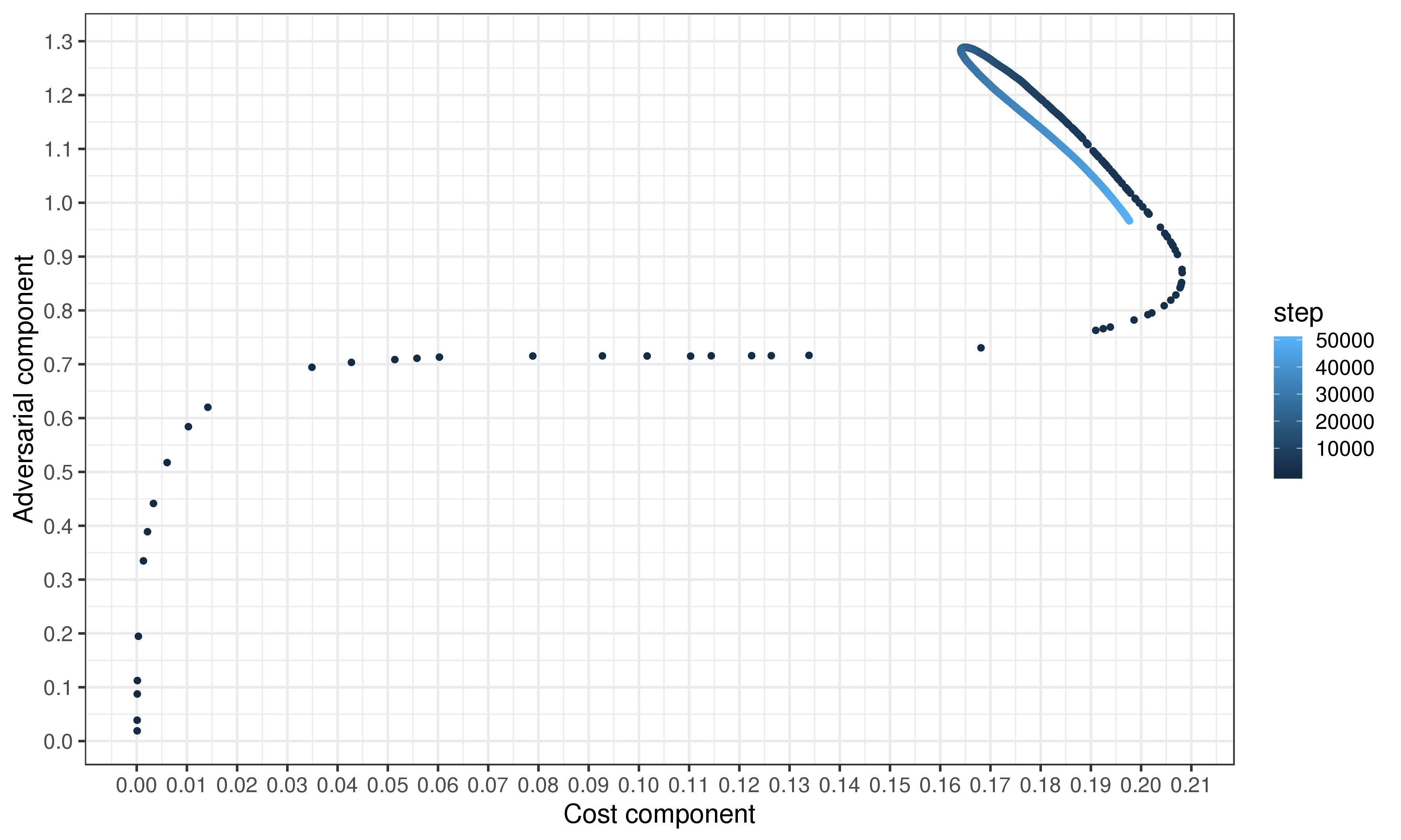}
    \end{figure}

 In Figure~\ref{fig:adversarial_cost_adv_losses} we compare
the adversarial and the cost component of the loss during the training.
We observe that at the beginning of the training (say $t<10k$) there
is a first phase displaying a quick increase in the
adversarial component which is then followed by a
saturation phase in which the adversarial component stays constant
while the cost component increases.
The first phase corresponds to the adversarial loss starting to
discriminate between $\mu$ and $\nu$ and the second phase to
the cost increasing as $\mu$ starts to move towards $\nu$.
\par In the training regime (say $t\ge 10k$ and $t<30k$) the
adversarial loss continues to grow driving a further
rearrangement of $T_{w(t)\#}\mu$ inside the support of $\nu$.
Finally, the last phase (say $t\ge 30k$) is dominated by a degree of
reduction of the adversarial loss leading to a collapse of $T_{w(t)\#}\mu$
inside the support of $\nu$.
This phenomenon points out that at a certain point the adversarial
network starts to fail at discriminating between $T_{w(t)\#}\mu$
and $\nu$.
Thus, for the rest of the paper we set the adversarial models aside.
 \begin{table}[ht]
\centering
\caption{Performance of ``adversarial'' models}\label{table:adv_models}
\begin{tabular}{r|r|r|r|r}
  \hline
model name & $\varepsilon_2$ & $\sigma(\varepsilon_2)$ &
  $t_{\textrm{min}}$ & $T$ \\
  \hline
adv\_l0.1\_2 & 0.70 & 0.010 & 22000 & 50000 \\
  adv\_l1\_10 & 0.49 & 0.013 & 4000 & 50000 \\
  adv\_l1\_2 & 0.37 & 0.037 & 39000 & 50000 \\
  adv\_l1\_2\_clip\_0.01 & 0.37 & 0.037 & 39000 & 50000 \\
  adv\_l10\_2 & 0.62 & 0.104 & 2000 & 50000 \\
  adv\_l100\_2 & 0.71 & 0.188 & 9000 & 50000 \\
   \hline
\end{tabular}
\end{table}

 \subsection{Comparison of Approaches based on duals \& supervised learning}
\label{subsec:dual_sup_comp}
\begin{figure}
    \centering
    \caption{Comparison of the final maps using the dual approach with
    $l2$-regularization;
    on the left using the mean aggregation and on the right the sum.}
                    \label{fig:l2_final_comp}
    \subfloat{\includegraphics[width=5cm]{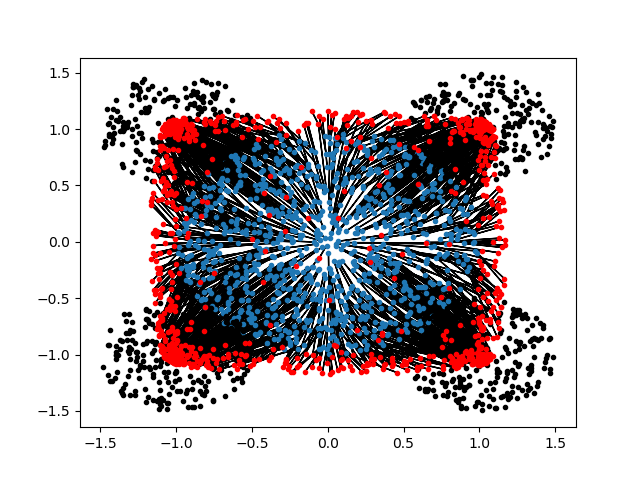}}
   \subfloat{\includegraphics[width=5cm]{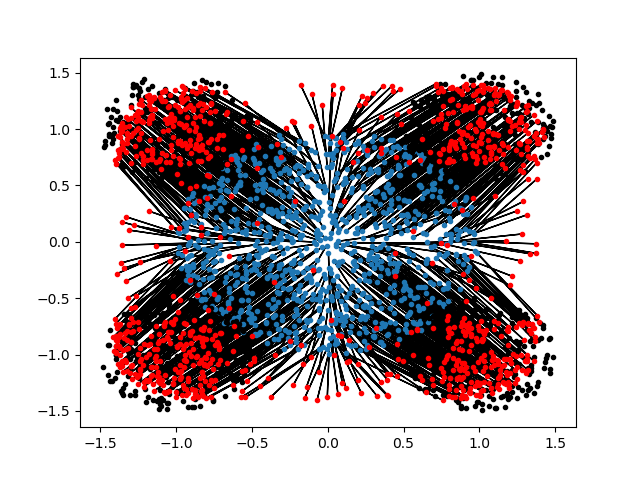}}
\end{figure}

\begin{figure}[ht]
        \caption{Convergence rate of models based on the dual approach.}
                        \label{fig:seguy_convergence}
        \includegraphics[height=7cm,width=12cm]{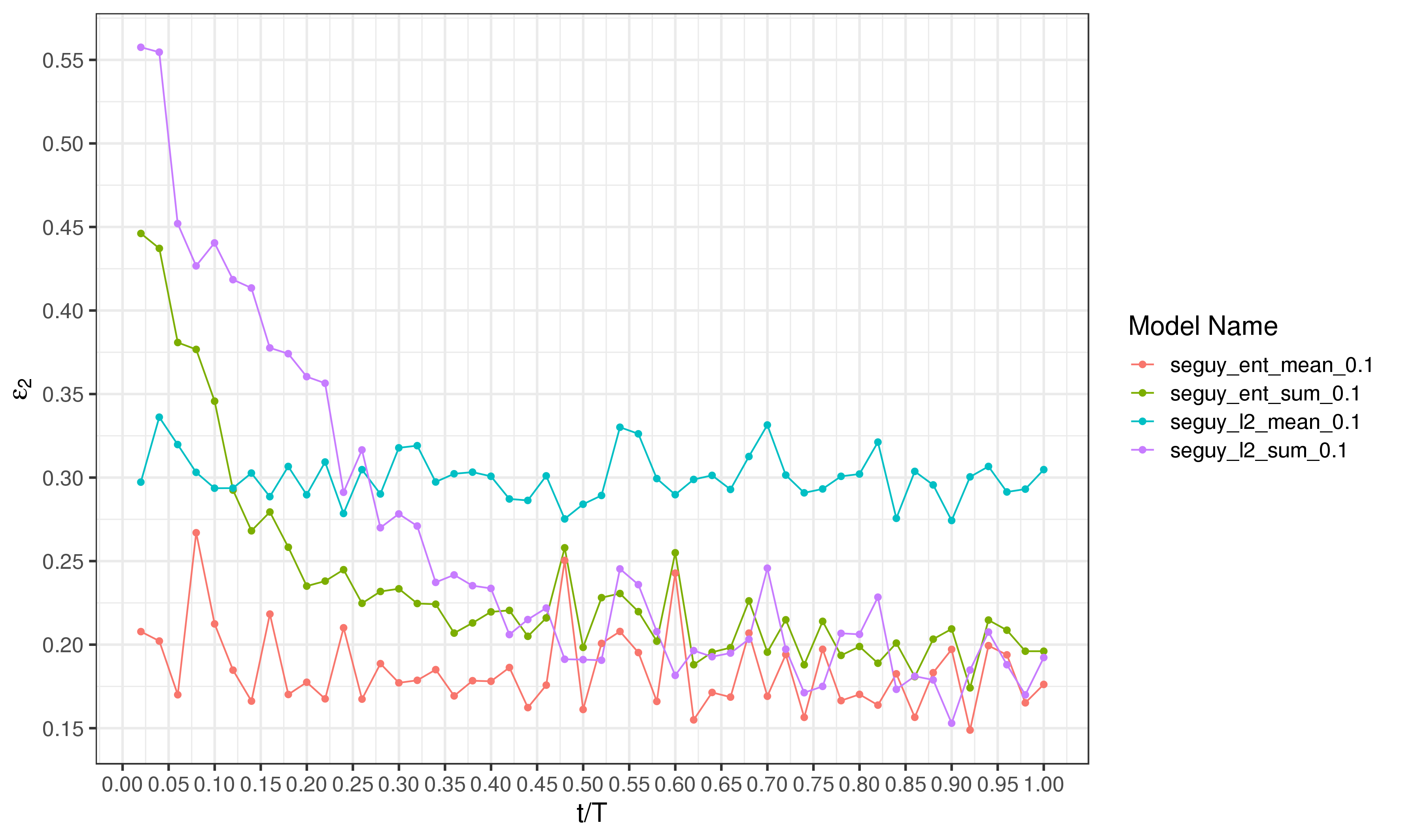}
    \end{figure}

\begin{figure}
    \centering
    \caption{Comparison of the final maps for \textit{seguy\_l2\_sum\_0.1}
    (left) and \textit{seguy\_ent\_mean\_0.1} (right).
    }
                    \label{fig:l2_ent_comp}
    \subfloat{\includegraphics[width=5cm]{l2-sum-final}}
   \subfloat{\includegraphics[width=5cm]{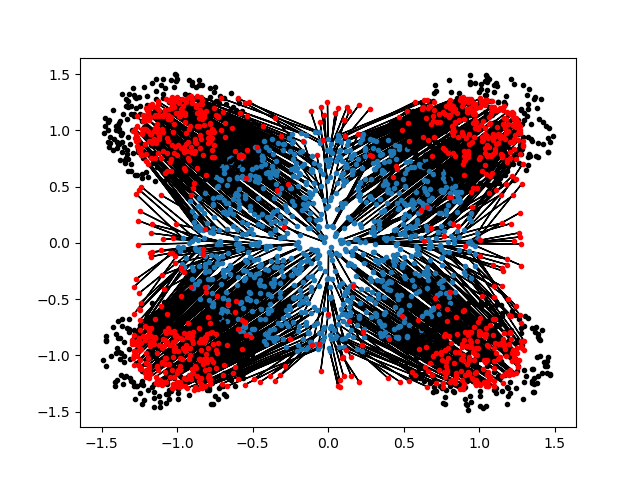}}
\end{figure}

    In table~\ref{table:dual_models} we can see a comparison
    of the performances of the algorithms based on the dual methods
    of Subsection~\ref{subsec:learn_tr_reg_dual_seguy} and
    supervised learning of Subsection~\ref{subsec:learn_tr_super_learn}.
    The models with a name \textit{seguy\_.*}  can use either the entropic
    or the $l_2$ regularization.
    Note that the name can contain either ``mean'' or ``sum'' depending on
    how the regularization is aggregated across the batch.
    In fact, we sum the potentials $\{u_i\}_{i=1}^B$, $\{v_j\}_{j=1}^B$
    on the batch;
    however, the regularization is a matrix $\{R_{i,j}\}_{i,j=1}^B$ and
    the mathematical formulation of duality~\cite[e.g.~Sec.~3]{genevay_large_opt} would suggest that we need
    to sum on one dimension (say $i$) and take the mean on the other one
(say $j$).
    However the equations in the original paper~\cite[Alg.~1]{seguy_neural_opt} suggest to use a
    double sum for the regularization term.
    While taking the sum degrades a bit the entropic model, it has a
    positive effect on the $l_2$-regularized model, which otherwise
    suffers from a collapsing phenomenon, see Figure~\ref{fig:l2_final_comp}.
    Overall \textit{seguy\_ent\_mean} and \textit{seguy\_l2\_sum} have
    comparable performances, see Figures~\ref{fig:seguy_convergence}
    and \ref{fig:l2_ent_comp},
    with the latter requiring less iterations.
\par
\begin{figure}[ht]
        \caption{Convergence rate of models based on the supervised.}
                        \label{fig:supervised_convergence}
        \includegraphics[height=7cm,width=14cm]{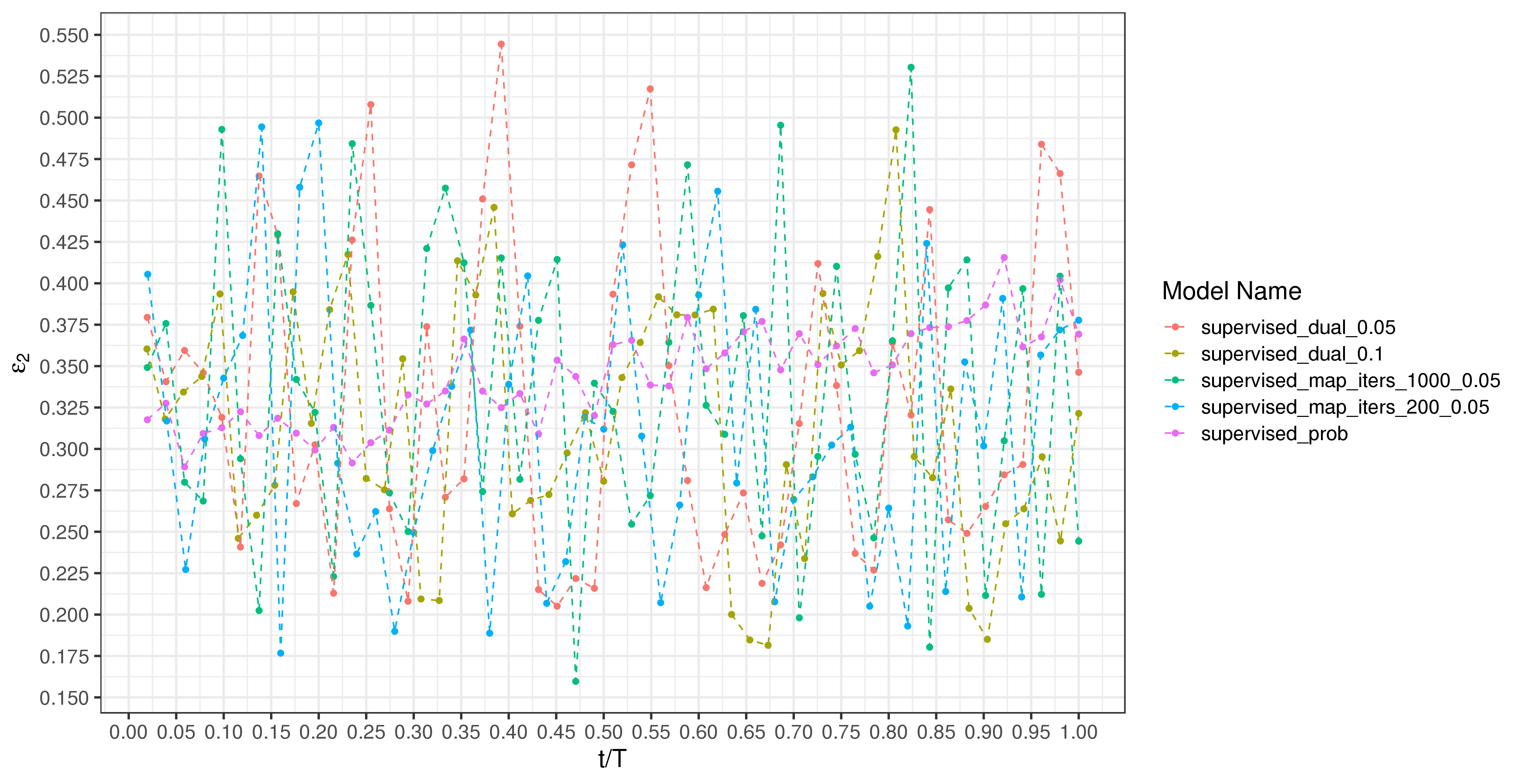}
    \end{figure}

 In Figure~\ref{fig:supervised_convergence} we can see the convergence
rate for the models using a supervised approach.
Except for \textit{supervised\_prob} all these models have a higher
variance $\sigma(\varepsilon_2)$ but, except for \textit{supervised\_prob},
the final performance is comparable to the one of
\textit{seguy\_ent\_mean} and \textit{seguy\_l2\_sum}.
\par
\begin{figure}
    \centering
    \caption{Poor quality of transport maps for \textit{supervised\_prob},
left at $t=1k$ and right at $t=47k$.}
                    \label{fig:poor_sup_prob}
    \subfloat{\includegraphics[width=5cm]{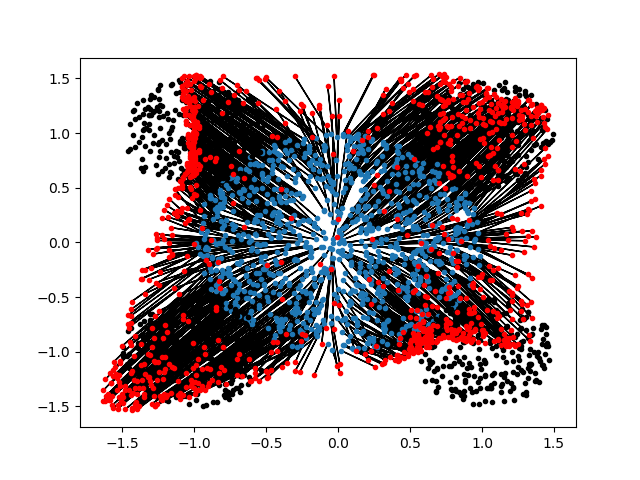}}
   \subfloat{\includegraphics[width=5cm]{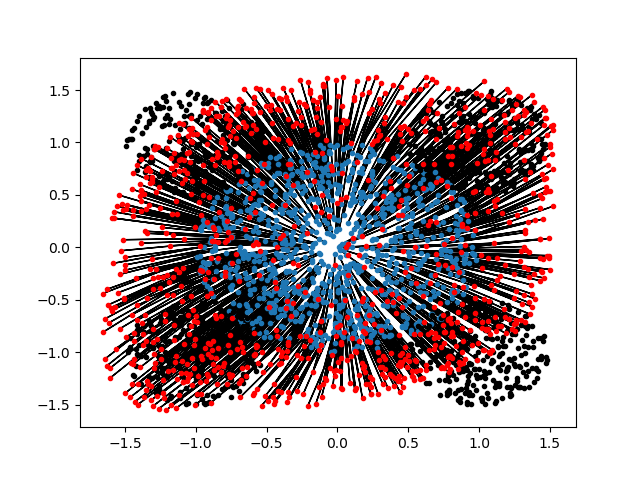}}
\end{figure}

 In the model \textit{supervised\_prob} we try to first learn
a transport plan and then use the heuristic~\eqref{eq:pi_to_T_heur} to
learn a transport map.
In practice this model performs poorly and the $\varepsilon_2$ starts to
grow;
on a qualitative inspection we observe that the first plans tend to distort the
geometry of $T_{\#}\mu$ relatively to that of $\nu$, while towards the end
of the training $T_{\#}\mu$ becomes too diffused, see Figure~\ref{fig:poor_sup_prob}.
Finally, we also point out that the training of this model is considerably slower
than for the others as it requires fitting a neural network for the plan
against a $B\times B$ matrix (while for potentials or maps we fit against
vectors of size $B$).
\par
\begin{figure}[ht]
    \caption{Comparison of maps learned with the supervised map
    approach.
    On the with relatively large $\varepsilon_2$ at $t=5k$
    and on the right the minimum $\varepsilon_2$ at $t=8k$.}
                    \label{fig:super_map_comp}
    \subfloat{\includegraphics[width=5cm]{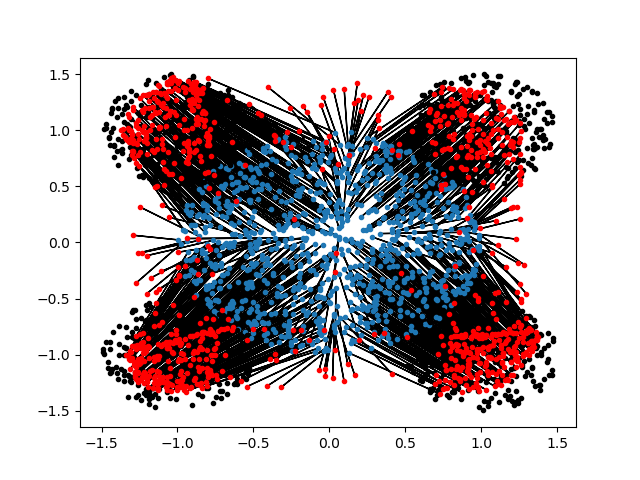}}
   \subfloat{\includegraphics[width=5cm]{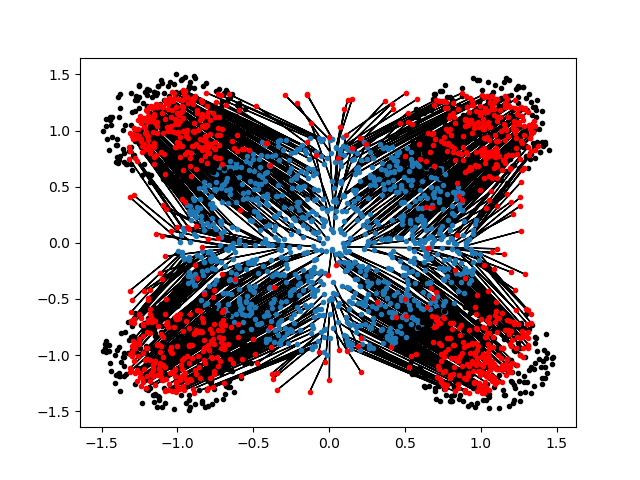}}
\end{figure}

 Both the methods based on the supervised dual or the supervised map
perform well.
The high variance seems to be an artifact of the iterations using supervised
data and can be reduced in practice by using a \emph{validation rule} which
decides when to stop the fitting of the map $T$.
Concretely, periodically one evaluates the performance of $T_{w(t)}$ against
a ``ground-truth'' $T_{\textrm{opt}}$ to decide when to stop the training of
$T$.
Note that even with the high variance the maps tend to stay qualitatively
closed to $T_{\textrm{opt}}$, i.e.~we do not observe phenomena like collapsing
or diffusion, see Figure~\ref{fig:super_map_comp}.
Finally, we point out that the approach based on directly learning $T$
is faster while both the models \textit{seguy\_.*} and \textit{supervised\_dual\_.*}
require a second step to fit $T$.
    \begin{table}[ht]
        \caption{Performance of models based on Regularized duals}\label{table:dual_models}

\centering
\begin{tabular}{r|r|r|r|r}
  \hline
  model name & $\varepsilon_2$ & $\sigma(\varepsilon_2)$ &
  $t_{\textrm{min}}$ & $T$ \\
  \hline
seguy\_ent\_mean\_0.1 & 0.15 & 0.021 & 9200 & 10000 \\
  seguy\_ent\_sum\_0.1 & 0.17 & 0.016 & 9200 & 10000 \\
  seguy\_l2\_mean\_0.1 & 0.27 & 0.012 & 4500 & 5000 \\
  seguy\_l2\_sum\_0.1 & 0.15 & 0.019 & 4500 & 5000 \\
  supervised\_dual\_0.05 & 0.21 & 0.093 & 13800 & 30600 \\
  supervised\_dual\_0.1 & 0.18 & 0.082 & 14000 & 20800 \\
  supervised\_map\_iters\_1000\_0.05 & 0.16 & 0.096 & 24000 & 51000 \\
  supervised\_map\_iters\_200\_0.05 & 0.18 & 0.084 & 8000 & 50000 \\
  supervised\_prob & 0.29 & 0.030 & 3000 & 51000 \\
   \hline
\end{tabular}
\end{table}
\subsection{Timings}
\label{subsec:timings}
We now discuss the time consumption for the models that we
find more promising, see Table~\ref{table:timings_models}.
We ran our experiments on a laptop with a $4$-Core Intel (R)
Core (TM) i5-5257U@2.70 GHz CPU. We used a docker image
of ubuntu:bionic and version 1.0 of PyTorch on Python 3.6.
\par In Table~\ref{table:timings_models} the seconds per step
are computed using the wall clock time.
The seconds per step are then used to estimate the time to $t_{\textrm{min}}$,
which we recall was the ``best'' of the iterations in the $S$ snapshots.
For models requiring two steps, namely training networks to optimize
a dual problem and use the potentials to learn a transport map,
we report two rows of timings, one for the dual and another for the map.
While \textit{secs per step} is an objective metric,
\textit{secs to $t_{\textrm{min}}$} must be taken with a
grain of salt as it depends on the parameters used to run
the corresponding script.
\par In general, total timings are relatively comparable.
However, we see the advantage of using models like
\textit{tp\_exp} or \textit{supervised\_map\_iters\_1000\_0.05}
that learn directly the transport map. Using Sinkhorn iterations
also gives a speed up compared to using the dual problem. For example
in \textit{seguy\_l2\_sum\_0.1} about $7.95\times 10^{-3}$ seconds
are required for each step of the dual problem, compared to the
$4.82\times 10^{-3}$ required by \textit{supervised\_dual\_0.1}.
    \begin{table}[ht]
        \caption{Timing statistics for models}\label{table:timings_models}

\centering
\begin{tabular}{r|r|r}
\hline
model name & secs per step & secs to $t_{\textrm{min}}$ \\
  \hline
tp\_exp & 0.00456 & 123.04 \\
  supervised\_map\_iters\_1000\_0.05 & 0.00303 & 72.60 \\
  seguy\_l2\_sum\_0.1 (dual) & 0.00795 & 35.78 \\
  seguy\_l2\_sum\_0.1 (map) & 0.01192 & 59.60 \\
  seguy\_ent\_mean\_0.1 (dual) & 0.00806 & 74.18 \\
  seguy\_ent\_mean\_0.1 (map) & 0.01245 & 124.48 \\
  supervised\_dual\_0.1 (dual) & 0.00482 & 67.48 \\
  supervised\_dual\_0.1 (map) & 0.00607 & 60.70 \\
   \hline
\end{tabular}
\end{table}
    \subsection{Conclusion}
\label{subsec:conclusion}
    We have compared a variety of approaches to find an optimal map
    between probability distributions.
    \par We find that, despite different theoretical/heuristic justifications,
    some algorithms yield similar good optimal maps.
    Specifically, we find flows using local Gaussian bumps, supervised
    learning approaches learning potentials or directly a transport map
    and the dual formulation of~\cite{seguy_neural_opt} to yield good results.
    In terms of time consumption, algorithms learning directly
    the transport map and using Sinkhorn's iterations are more favorable.
    \par On the other hand, we also find other approaches to under-perform
    or being unstable.
    In particular, flows using the covariance loss or the \textbf{discr @ } $N$
    seem to yield poor maps.
    Approaches using adversarial training yield poor maps and are
    also unstable to train.
    \bibliographystyle{alpha}
    \bibliography{the-biblio}

\newcommand{\etalchar}[1]{$^{#1}$}
\begin{thebibliography}{GPAM{\etalchar{+}}14}

\bibitem[AB17]{arjovsky_principled}
Martin {Arjovsky} and L{\'e}on {Bottou}.
\newblock {Towards Principled Methods for Training Generative Adversarial
  Networks}.
\newblock {\em arXiv e-prints}, page arXiv:1701.04862, Jan 2017.

\bibitem[ACB17]{arjovsky_wgan}
Martin {Arjovsky}, Soumith {Chintala}, and L{\'e}on {Bottou}.
\newblock {Wasserstein GAN}.
\newblock {\em arXiv e-prints}, page arXiv:1701.07875, Jan 2017.

\bibitem[Cut13]{cuturi_lightspeed}
Marco Cuturi.
\newblock Sinkhorn distances: Lightspeed computation of optimal transport.
\newblock In {\em Proceedings of the 26th International Conference on Neural
  Information Processing Systems - Volume 2}, NIPS'13, pages 2292--2300, USA,
  2013. Curran Associates Inc.

\bibitem[GCPB16]{genevay_large_opt}
Aude Genevay, Marco Cuturi, Gabriel Peyr{\'e}, and Francis Bach.
\newblock Stochastic optimization for large-scale optimal transport.
\newblock In {\em Proceedings of the 30th International Conference on Neural
  Information Processing Systems}, NIPS'16, pages 3440--3448, USA, 2016. Curran
  Associates Inc.

\bibitem[GM17]{maggioni_package}
Samuel Gerber and Mauro Maggioni.
\newblock Multiscale strategies for computing optimal transport.
\newblock {\em CoRR}, abs/1708.02469, 2017.

\bibitem[GPAM{\etalchar{+}}14]{goodfellow_gans}
Ian Goodfellow, Jean Pouget-Abadie, Mehdi Mirza, Bing Xu, David Warde-Farley,
  Sherjil Ozair, Aaron Courville, and Yoshua Bengio.
\newblock Generative adversarial nets.
\newblock In Z.~Ghahramani, M.~Welling, C.~Cortes, N.~D. Lawrence, and K.~Q.
  Weinberger, editors, {\em Advances in Neural Information Processing Systems
  27}, pages 2672--2680. Curran Associates, Inc., 2014.

\bibitem[KS12]{quasi_experiments_video_streaming}
S.~Shunmuga Krishnan and Ramesh~K. Sitaraman.
\newblock Video stream quality impacts viewer behavior: Inferring causality
  using quasi-experimental designs.
\newblock In {\em Proceedings of the 2012 Internet Measurement Conference}, IMC
  '12, pages 211--224, New York, NY, USA, 2012. ACM.

\bibitem[KSKW15]{kusner-embeddings}
Matt Kusner, Yu~Sun, Nicholas Kolkin, and Kilian Weinberger.
\newblock From word embeddings to document distances.
\newblock In Francis Bach and David Blei, editors, {\em Proceedings of the 32nd
  International Conference on Machine Learning}, volume~37 of {\em Proceedings
  of Machine Learning Research}, pages 957--966, Lille, France, 07--09 Jul
  2015. PMLR.

\bibitem[LA08]{ambrosio_book}
Giuseppe~Savare Luigi~Ambrosio, Nicola~Giglio.
\newblock {\em Gradient Flows In Metric Spaces and in the Space of Probability
  Measures}.
\newblock Birkhäuser Basel, 2008.
\newblock Available at: http://www2.stat.duke.edu/~sayan/ambrosio.pdf.

\bibitem[LV09]{lott_rcd}
John Lott and Cédric Villani.
\newblock Ricci curvature for metric-measure spaces via optimal transport.
\newblock {\em Annals of Mathematics}, 169(3):903--991, 2009.

\bibitem[Net]{netflix_quasi_experiments}
Netflix.
\newblock Quasi experimentation at netflix.
\newblock
  \url{https://medium.com/netflix-techblog/quasi-experimentation-at-netflix-566b57d2e362}.
\newblock Technology Blog (Sep 2018), Accessed at: 2019-07-30.

\bibitem[OZM{\etalchar{+}}16]{orlova_medical}
Darya~Y. Orlova, Noah Zimmerman, Stephen Meehan, Connor Meehan, Jeffrey Waters,
  Eliver E.~B. Ghosn, Alexander Filatenkov, Gleb~A. Kolyagin, Yael Gernez,
  Shanel Tsuda, Wayne Moore, Richard~B. Moss, Leonore~A. Herzenberg, and
  Guenther Walther.
\newblock Earth mover’s distance (emd): A true metric for comparing biomarker
  expression levels in cell populations.
\newblock {\em PLOS ONE}, 11(3):1--14, 03 2016.

\bibitem[PC19]{cuturi_book}
Gabriel Peyré and Marco Cuturi.
\newblock Computational optimal transport.
\newblock {\em Foundations and Trends® in Machine Learning}, 11(5-6):355--607,
  2019.

\bibitem[PvF{\etalchar{+}}18]{patrini_sink_auto}
Giorgio {Patrini}, Rianne {van den Berg}, Patrick {Forr{\'e}}, Marcello
  {Carioni}, Samarth {Bhargav}, Max {Welling}, Tim {Genewein}, and Frank
  {Nielsen}.
\newblock {Sinkhorn AutoEncoders}.
\newblock {\em arXiv e-prints}, page arXiv:1810.01118, Oct 2018.

\bibitem[San15]{santambrogio_book}
Filippo Santambrogio.
\newblock {\em Optimal Transport for Applied Mathematicians}.
\newblock Birkhäuser Basel, 2015.
\newblock Available at:
  https://www.math.u-psud.fr/$\sim$filippo/OTAM-cvgmt.pdf.

\bibitem[SDF{\etalchar{+}}18]{seguy_neural_opt}
Vivien Seguy, Bharath~Bhushan Damodaran, R{\'e}mi Flamary, Nicolas Courty,
  Antoine Rolet, and Mathieu Blondel.
\newblock Large-scale optimal transport and mapping estimation.
\newblock In {\em Proceedings of the International Conference in Learning
  Representations}, 2018.

\bibitem[SdGP{\etalchar{+}}15]{solomon_conv_wasserstein}
Justin Solomon, Fernando de~Goes, Gabriel Peyr{\'e}, Marco Cuturi, Adrian
  Butscher, Andy Nguyen, Tao Du, and Leonidas Guibas.
\newblock Convolutional wasserstein distances: Efficient optimal transportation
  on geometric domains.
\newblock {\em ACM Trans. Graph.}, 34(4):66:1--66:11, July 2015.

\bibitem[SK67]{sinkhorn1967}
Richard Sinkhorn and Paul Knopp.
\newblock Concerning nonnegative matrices and doubly stochastic matrices.
\newblock {\em Pacific J. Math.}, 21(2):343--348, 1967.

\bibitem[TT16]{trigila_thesis}
Giulio Trigila and Esteban~G. Tabak.
\newblock Data-driven optimal transport.
\newblock {\em Communications on Pure and Applied Mathematics}, 69(4):613--648,
  2016.

\bibitem[WB17]{weed_convergence}
Jonathan {Weed} and Francis {Bach}.
\newblock {Sharp asymptotic and finite-sample rates of convergence of empirical
  measures in Wasserstein distance}.
\newblock {\em arXiv e-prints}, page arXiv:1707.00087, Jun 2017.

\bibitem[ZML16]{lecun_energy_wgans}
Junbo {Zhao}, Michael {Mathieu}, and Yann {LeCun}.
\newblock {Energy-based Generative Adversarial Network}.
\newblock {\em arXiv e-prints}, page arXiv:1609.03126, Sep 2016.

\end{thebibliography}
\end{document}